\documentclass{article}

\usepackage{PRIMEarxiv}

\usepackage[utf8]{inputenc} %
\usepackage[T1]{fontenc}    %

\usepackage{hyperref}       %
\hypersetup{pdftitle={Let the Gate Decide: Per-Lookup Precision for Quantized Conditional Memory in LLMs},
  pdfauthor={Tianqi Chen, Xunchao Li, Linchuan Xie, Guanghua Yu, Jianchen Zhu, Peisong Wang, Jian Cheng}}
\usepackage{url}            %
\usepackage{booktabs}       %
\usepackage{amsfonts}       %
\usepackage{nicefrac}       %
\usepackage{microtype}      %
\usepackage{graphicx}
\usepackage[numbers,sort&compress]{natbib}

\usepackage{amsmath,amsfonts,bm}

\def\eqref#1{equation~\ref{#1}}

\def\1{\bm{1}}

\DeclareMathAlphabet{\mathsfit}{\encodingdefault}{\sfdefault}{m}{sl}
\SetMathAlphabet{\mathsfit}{bold}{\encodingdefault}{\sfdefault}{bx}{n}

\usepackage{multirow}
\usepackage{amsmath,amssymb}
\usepackage{amsthm}
\usepackage{algorithm}
\usepackage{algpseudocode}

\renewcommand{\eqref}[1]{Equation~(\ref{#1})}

\usepackage{xcolor}
\usepackage{colortbl}
\usepackage{arydshln}
\usepackage{subcaption}
\usepackage{wrapfig}
\usepackage{needspace}
\usepackage{placeins}  %

\definecolor{oursrowtint}{RGB}{234,239,249}
\newcommand{\ours}{\rowcolor{oursrowtint}}
\newcommand{\method}{GateRQ}
\newcommand{\accfigwidth}{0.85\textwidth}

\newcommand{\todocell}[1]{\textcolor{gray}{--}}
\newcommand{\todoinline}[1]{\textcolor{red}{[TODO: #1]}}

\title{Let the Gate Decide: Per-Lookup Precision\\
for Quantized Conditional Memory in LLMs}

\author{
  \textbf{Tianqi Chen$^{1,3}$\quad Xunchao Li$^{2}$\quad Linchuan Xie$^{2}$\quad Guanghua Yu$^{2}$} \\
  \textbf{Jianchen Zhu$^{2}$\quad Peisong Wang$^{1,3}$\quad Jian Cheng$^{1,3}$\,$^{\dagger}$} \\[6pt]
  $^{1}$C$^{2}$DL, Institute of Automation, Chinese Academy of Sciences
  \quad $^{2}$Tencent \\
  $^{3}$School of Artificial Intelligence, University of Chinese Academy of Sciences \\[4pt]
  \texttt{chentianqi2023@ia.ac.cn}, \texttt{lucayu@tencent.com},
  \texttt{jcheng@nlpr.ia.ac.cn} \\[2pt]
  $^{\dagger}$Corresponding author.
}

\begin{document}

\maketitle

\begin{abstract}
Conditional memory layers, such as DeepSeek's Engram, attach large hashed
$n$-gram tables to a language model, and removing them from
DeepSeek-V4.1-Flash costs $36.8$ PopQA points. Low-bit quantization loses
long-tail facts first, and the standard remedy, giving more bits to frequently
accessed rows, backfires: refining lookups by access frequency is worse than
refining them at random. We trace this to the gate that scales each retrieved
row. A lookup's quantization error is weighted by its gate, which is small for
frequent $n$-grams, large for the rare $n$-grams of long-tail entities, and
computable from a $1$-bit code. \method{} therefore stores each row as a
ladder of $1$-bit residual codes, keeps level~1 in GPU memory to compute the
gate, and reads deeper levels only for the lookups the gate ranks highest.
Without retraining, \method{} cuts GPU table memory by $7.8\times$ on
DeepSeek-V4.1-Flash, which raises vLLM throughput by $11\%$--$31\%$ over FP8
tables, while losing about $1$ PopQA point at $1.50$ bits read per lookup and
$0.2$ points at $2.10$ bits across two models.
\end{abstract}


\section{Introduction}
\label{sec:intro}

Recent large language models (LLMs) place part of their parameters in
\emph{conditional memory}: hash tables, keyed by the preceding $n$-gram,
that sit beside the Transformer backbone, as in DeepSeek's Engram
\citep{engram} and Qwen's Per-Layer Embedding (PLE)
\citep{qwen38next}. Each token retrieves a fixed number of rows,
and a gate computed from the current hidden state scales the retrieved
vectors before they enter the residual stream. The tables thus add many
parameters at little compute cost, and they hold much of the model's factual
knowledge. Zeroing them in DeepSeek-V4.1-Flash costs $36.8$ points on PopQA
\citep{popqa}, a benchmark of questions about long-tail entities, more than
two thirds of its accuracy; the answers stay well formed, but the facts are
wrong. The tables are also large: $203$~GB ($189$~GiB) in FP8, $40\%$ of the
checkpoint. On a node of $8$ GPUs, keeping them in GPU memory
takes $23.6$~GiB on every GPU and leaves a KV cache nine times smaller (as
reported by vLLM) than when they are offloaded; offloading them instead pins
all $189$~GiB in host memory.

Much of what these tables contribute is long-tail knowledge: facts about
rare entities, such as the author of a little-known novel, that the backbone
does not recover on its own. Quantization, the natural way to shrink the
tables, erases exactly these facts first (Section~\ref{sec:motivation}), so a
compression budget should not be spread evenly across rows. Recommender
systems, whose embedding tables face the same pressure, have a standard rule
for deciding which rows to keep precise: the frequently accessed ones, which
receive larger dimensions \citep{ginart2021mixed} or higher precision
\citep{yang2020mixed,li2024embedding}. The rule rests on an assumption: every
access contributes equally to the output, so frequent rows, read most often,
accumulate the most quantization error.

This assumption does not hold for conditional memory. Each retrieved row
passes through a gate before it enters the residual stream, so its
quantization error is paid at every lookup, scaled by that lookup's gate
(Section~\ref{sec:lookup-distortion}). The gates, moreover, run against
frequency: frequent $n$-grams are mostly function words and common phrases,
which receive small gates, whereas the rare $n$-grams of named entities
receive large ones. In the question ``Who is the author of
\emph{The Pianoplayers}?'', the gate is smallest at the token \texttt{The}
and largest on the three tokens of \texttt{Pianoplayers}
(Figure~\ref{fig:motivation}a). Over PopQA questions, rows seen more than a
thousand times serve a third of the $2$-gram lookups of block~2 but, under
uniform quantization, carry under $4\%$ of the output error
(Figure~\ref{fig:motivation}b). A
frequency rule therefore spends its bits where they matter least: choosing
lookups to refine by frequency does worse than choosing them at random
(Sections~\ref{sec:observations} and~\ref{sec:results-main}).

One might hope to repair the rule instead, ranking rows by their expected
gate rather than by frequency. This helps, but it still misses many
important lookups, for two reasons. The rows that matter most belong to rare
$n$-grams, which a counting corpus sees too seldom to estimate their gate;
and even for common rows, the gate changes with context and can be large in
one sentence and near zero in another (Section~\ref{sec:observations}).
Precision should therefore be decided per lookup, from the gate at hand.
This seems to require the very precision one wants to save, since the gate
is computed from the retrieved rows. It does not: the gate is insensitive to
quantization, and gates computed from a $1$-bit code closely match those
computed from the FP8 table ($r\ge0.989$; Section~\ref{sec:observations}).
Whether a lookup needs higher precision can thus be decided before any
high-precision data is read.

\method{} (gate-driven residual quantization;
Figure~\ref{fig:method-overview}) combines two pieces. The first is a code
whose precision can be raised after a row has been read: each row is stored
as a ladder of $1$-bit residual codes, with level~1 in GPU memory and deeper
levels in host memory. The second is a rule that decides, per lookup, how far
up the ladder to go: every lookup reads level~1 to obtain its key and gate,
and under a fixed budget of bits per lookup, the lookups that the gate marks
as most important fetch deeper levels, which refine only the value. Precision
thus becomes a property of each access rather than of the stored row.

Our contributions are as follows:
\begin{itemize}
\item \textbf{Analysis.} We show three properties of quantization error in
gated conditional memory: the gate error is negligible, so the gate can be
computed accurately from a $1$-bit code; importance runs against access
frequency, so frequency-based allocation fails; and importance is concentrated
in a few lookups whose identity depends on context, so row-level statistics
cannot find them (Section~\ref{sec:motivation}).
\item \textbf{Method.} We propose \method{}, a post-training scheme that
stores each row as a ladder of $1$-bit residual codes and uses the gate
computed from the first level to choose how many levels each lookup reads.
It needs no retraining, keeps the model's gating mechanism intact, and applies
to any codec that can encode residuals (Section~\ref{sec:method}).
\item \textbf{Evaluation.} We evaluate on two models, three codecs
(trellis-coded quantization, TCQ; integer, INT; product quantization, PQ), and
two read budgets ($12$ settings). At the same number of bits read per lookup,
gate-based selection beats selection by expected gate, at random, and by
access frequency in all $36$ paired PopQA comparisons. With TCQ
\citep{tcq,qtip}, reading $1.50$ and $2.10$ bits per lookup costs about $1$ and
$0.2$ PopQA points relative to FP8 and is statistically indistinguishable from
reading the full two- and three-level ladders, which take $0.63$--$1.20$ more
bits per lookup (Section~\ref{sec:results-main}).
\item \textbf{System.} In vLLM \citep{pagedattention} on $8$ GPUs,
\method{} cuts the
table memory from $23.6$ to $3.0$~GiB per GPU, which enlarges the KV cache
$8.1\times$ (as reported by vLLM) and raises throughput by $11\%$--$31\%$ over
FP8 tables in GPU memory. Against FP8 host offloading, it matches throughput
(within $1.5\%$) with $26$--$49$~GiB of pinned host memory instead of
$189$~GiB (Appendix~\ref{app:serving}).
\end{itemize}

\section{Background: Conditional Memory Layers}
\label{sec:background}
\label{sec:prelim}

We first describe conditional memory following Engram \citep[Section~2]{engram},
then distinguish the implementations in the two models studied here.

\paragraph{Hashed $n$-gram retrieval.}
Given a token sequence $X=(x_1,\ldots,x_T)$, Engram first compresses the
vocabulary. A precomputed surjection $\mathcal P$ maps tokens with the same
normalized text, such as case variants, to canonical IDs
$x'_t=\mathcal P(x_t)$. At position $t$, it forms suffix $n$-grams
$g_{t,n}=(x'_{t-n+1},\ldots,x'_t)$ for $n=2,\ldots,N$. Since the space of
all such $n$-grams is too large to parameterize directly, each order uses
$K$ hash heads to address tables and mitigate collisions. Head $k$ uses a
multiplicative--XOR hash $\varphi_{n,k}$ to select a row in a table
$\mathbf E_{n,k}$ of prime size $p_{n,k}$:
\begin{equation}
z_{t,n,k}=\varphi_{n,k}(g_{t,n}),\qquad
\mathbf e_{t,n,k}=\mathbf E_{n,k}[z_{t,n,k}]\in\mathbb R^{d_e}.
\label{eq:engram-lookup}
\end{equation}
The $(N-1)K$ retrieved vectors are concatenated into a memory vector
$\mathbf e_t$. We combine $(n,k)$ into a head index $j=1,\ldots,J$, where
$J=(N-1)K$, and write the row retrieved by head $j$ as $\mathbf e_{t,j}$.
One row access $(t,j)$ is a \emph{lookup}. Its row ID, denoted $r(t,j)$,
depends only on tokens, not on the hidden state.

\paragraph{Context-dependent gating.}
Retrieved vectors do not themselves depend on the current hidden state and
may be noisy because of collisions or ambiguity. Engram uses the hidden
state as a query and the memory vector as the source of keys and values to
compute a scalar gate. With a multi-branch backbone, the residual stream has
$M$ branches ($M=4$ in both models). All branches share the table and value
projection $\mathbf W_V$, while branch $m$ has its own key projection
$\mathbf W_K^{(m)}$:
\begin{equation}
\mathbf k_t^{(m)}=\mathbf W_K^{(m)}\mathbf e_t,\qquad
\mathbf v_t=\mathbf W_V\mathbf e_t
=\sum_{j=1}^{J}\mathbf W_V^{(j)}\mathbf e_{t,j}.
\label{eq:engram-kv}
\end{equation}
Writing $\bar{\mathbf h}_t^{(m)}=\operatorname{RMSNorm}(\mathbf h_t^{(m)})$
and $\bar{\mathbf k}_t^{(m)}=\operatorname{RMSNorm}(\mathbf k_t^{(m)})$,
the original Engram gate and gated value are
\begin{equation}
\alpha_t^{(m)}=\sigma\!\left(
\frac{\bar{\mathbf h}_t^{(m)\top}\bar{\mathbf k}_t^{(m)}}{\sqrt d}\right),
\qquad \mathbf u_t^{(m)}=\alpha_t^{(m)}\mathbf v_t.
\label{eq:engram}
\end{equation}
Here $\mathbf h_t^{(m)}\in\mathbb R^d$ is the input state of branch $m$,
$\mathbf W_V^{(j)}$ is the column block of $\mathbf W_V$ corresponding to
head $j$, and $\sigma$ is sigmoid. When memory and context disagree, the gate
suppresses the retrieved contribution. Engram's visualizations show gates
opening for local patterns such as multi-token named entities and fixed
phrases \citep[Figure~7]{engram}.

In the original Engram layer, the gated values subsequently pass through a
short depthwise causal convolution. If $\tilde{\mathbf V}$ denotes the
sequence of gated values, the output is
\[
\mathbf Y=\operatorname{SiLU}\!\left(
\operatorname{Conv1D}(\operatorname{RMSNorm}(\tilde{\mathbf V}))\right)
+\tilde{\mathbf V},
\]
with kernel width $4$ and dilation $N$. This output is added to the backbone
before the block's attention and MoE. A retrieved row thus affects the layer
through two paths: the key sets the scalar gate, and the value supplies the
content written into the residual stream.

\paragraph{The two model implementations.}
DeepSeek-V4.1-Flash has Engram layers in blocks~2 and~15, as in Engram-27B.
It uses $N=4$, $K=8$, hence $J=24$, with $d_e=256$. Vocabulary compression
maps $129{,}280$ token IDs to $99{,}092$ canonical IDs; its backbone uses
four-branch manifold-constrained hyper-connections (mHC)
\citep{xie2025mhc}. Qwen3.8-Flash-Next \citep{qwen38next} has one PLE layer
in block~2, before that block's Gated DeltaNet. It uses $N=3$, $K=8$,
$J=16$, and $d_e=160$, hashes the original token IDs without vocabulary
compression, and has a four-branch gated residual backbone.
Both implementations apply a signed square root,
$\operatorname{sign}(x)\sqrt{|x|}$, to the scaled dot product before sigmoid
in \eqref{eq:engram}. DeepSeek removes the short convolution and adds the
gated value directly to each branch; Qwen retains a width-$4$, dilation-$3$
convolution. Table sizes and storage formats are given in Appendix~\ref{app:models}.

\section{Where Quantization Loss Comes From}
\label{sec:motivation}

\begin{figure}[t]
\centering
\includegraphics[width=\textwidth]{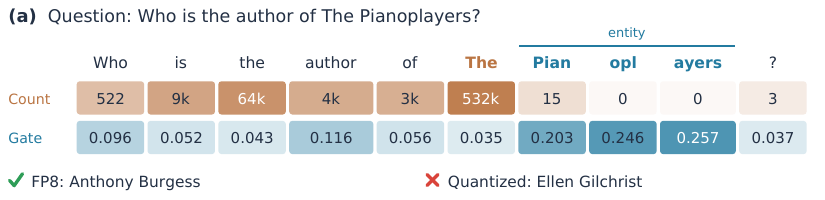}\\[2pt]
\includegraphics[width=0.486\textwidth]{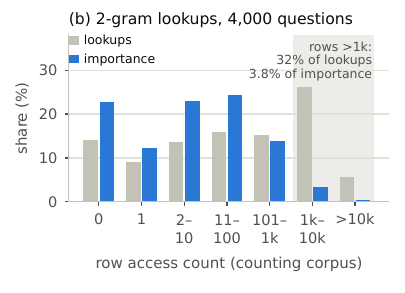}\hfill
\includegraphics[width=0.504\textwidth]{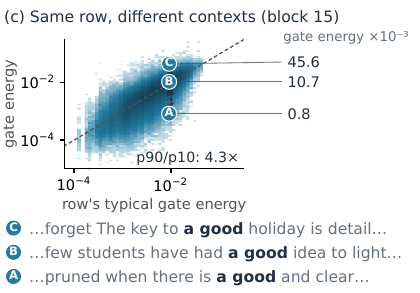}
\caption{\textbf{Importance follows the gate, not the access count}
(DeepSeek-V4.1-Flash).
(a)~A PopQA question that FP8 answers correctly and the quantized tables do
not: the entity carries most of the gate, frequent function words little.
(b)~Rows are binned offline by access count; the lookups and importance
(\eqref{eq:distortion}) of $4{,}000$ PopQA questions are then tallied per bin
(eight $2$-gram heads of block~2). Frequent rows (over $1{,}000$ accesses)
serve a third of the lookups but carry under $4\%$ of the importance.
(c)~Average gate energy of a row ($x$; geometric mean over its lookups)
against the gate energy of its individual lookups ($y$), for rows looked up at
least $20$ times (block~15, RACE passages);
dark dots: the $288$ lookups of the row read for \emph{a good}, A--C its
lowest, median and highest; p90/p10: median ratio of a row's 90th- to
10th-percentile gate energy.}
\label{fig:motivation}
\end{figure}

Quantizing the tables to one bit with TCQ, which gives level~1 of the
ladder in Section~\ref{sec:method}, costs DeepSeek-V4.1-Flash $3.87$ PopQA
points, and the cost is uneven: $6.24$ points on the least popular quartile
of entities, but only $1.51$ on the most popular (Appendix~\ref{app:popularity}). This
section asks which lookups the loss comes from, why access frequency fails
to find them, and whether they can be found before any high-precision data
is read. Section~\ref{sec:lookup-distortion} writes the distortion of a
layer as a sum over lookups. Section~\ref{sec:observations} then makes three
observations, from records of the gate and row of every lookup that the FP8
model makes on $4{,}000$ PopQA questions and on English text (held-out
documents and RACE passages; Figure~\ref{fig:motivation}).

\subsection{Distortion of a Lookup}
\label{sec:lookup-distortion}
\label{sec:method-prelim}

The quantization error
$\boldsymbol\delta_{t,j}=\hat{\mathbf e}_{t,j}-\mathbf e_{t,j}$ of a retrieved
row reaches the residual stream through both paths of
Section~\ref{sec:background}: it shifts the gate of branch $m$ by
$\Delta\alpha_t^{(m)}$ and the value by
$\Delta\mathbf v_t=\sum_j\mathbf W_V^{(j)}\boldsymbol\delta_{t,j}$, so the
gated update changes by
\begin{equation}
\Delta\mathbf u_t^{(m)}
=\hat\alpha_t^{(m)}\hat{\mathbf v}_t-\alpha_t^{(m)}\mathbf v_t
=\underbrace{\hat\alpha_t^{(m)}\Delta\mathbf v_t}_{\text{value error}}
+\underbrace{\Delta\alpha_t^{(m)}\mathbf v_t}_{\text{gate error}}.
\label{eq:error-split}
\end{equation}
The gate error is negligible (Observation~1), so we keep only the value
error, with $\hat\alpha_t^{(m)}\approx\alpha_t^{(m)}$. Treating the errors of
different lookups as uncorrelated and isotropic (Appendix~\ref{app:assumptions}), its expected
energy separates over lookups:
\begin{equation}
\mathbb E\sum_{t,m}\big\|\alpha_t^{(m)}\Delta\mathbf v_t\big\|_2^2
\approx\sum_{t,j}I_{t,j}\,\varepsilon_{t,j},
\qquad I_{t,j}=G_t\,\tau_j\,\|\mathbf e_{t,j}\|_2^2,
\label{eq:distortion}
\end{equation}
where
$\varepsilon_{t,j}=\|\boldsymbol\delta_{t,j}\|_2^2/\|\mathbf e_{t,j}\|_2^2$
is the relative error of the lookup, $G_t=\sum_m(\alpha_t^{(m)})^2$ is the
\emph{gate energy} at position~$t$, and
$\tau_j=\operatorname{tr}(\mathbf W_V^{(j)\top}\mathbf W_V^{(j)})/d_e$ is
the gain of head~$j$. We call $I_{t,j}$ the \emph{importance} of the lookup.
Its head gain and row energy are fixed for a given head and row, but its gate
energy depends on the context of each access. \eqref{eq:distortion} is a
local proxy rather than a task loss (for Qwen it also ignores the short
convolution); Section~\ref{sec:results} tests it end to end.

\subsection{Three Observations}
\label{sec:observations}

\setlength{\columnsep}{14pt}
\begin{wraptable}{r}{0.44\textwidth}
\vspace{-0.1em}
\centering
\captionsetup{font=small,justification=raggedright}
\caption{\textbf{Where the loss comes from} (DeepSeek-V4.1-Flash). PopQA
accuracy change (points) from FP8 ($52.39$) when only the gates, only the
values, or both use the quantized table.}
\label{tab:kvsplit}
\small
\setlength{\tabcolsep}{6pt}
\renewcommand{\arraystretch}{1.3}
\begin{tabular*}{\linewidth}{@{\extracolsep{\fill}}lccc}
\toprule
 & \multicolumn{3}{c}{Quantized} \\
\cmidrule(lr){2-4}
Bits & Gates & Values & Both \\
\midrule
1.06 & $+0.08$ & $-4.03$ & $-3.87$ \\
2.13 & $+0.15$ & $-0.81$ & $-0.88$ \\
3.19 & $+0.13$ & $-0.19$ & $-0.15$ \\
\bottomrule
\end{tabular*}
\vspace{-1.0em}
\end{wraptable}

\paragraph{Observation 1: The gate error is negligible; the loss comes from
the value.}
\label{par:stable-gates}
The gate error in \eqref{eq:error-split} is small because quantization barely
moves the gate. In every memory layer of both models, gates computed from the
1-bit table correlate with those computed from the FP8 table at $0.989$ to
$1.000$ position by position, and the two gates agree on which lookups matter
most: the top $10\%$ of DeepSeek's lookups by importance under each gate
overlap by $98.1\%$ in block~2 and $92.8\%$ in block~15 (held-out English
text; Appendix~\ref{app:gate}). Accuracy agrees (Table~\ref{tab:kvsplit}). Computing
only the gates from the quantized table leaves PopQA at FP8's level at every
depth, costing at most $0.22$ points at $95\%$ confidence (Appendix~\ref{app:gate}),
whereas quantizing only the values costs as much as quantizing both.

\paragraph{Observation 2: Access frequency runs against importance.}
With the gate error set aside, the cost of quantizing a lookup scales with
its importance $I_{t,j}$ in \eqref{eq:distortion}. Access frequency, the
usual proxy for how much a row matters, points the opposite way. In the
PopQA question of Figure~\ref{fig:motivation}a, which FP8 answers correctly
and the 1-bit table does not, the $2$-gram rows read at \texttt{The} were seen
about half a million times in our counting corpus ($115$M RedPajama tokens),
yet the gate there is only $0.035$; the rows read at the three tokens of
\texttt{Pianoplayers} were seen at most $15$ times, yet receive the
largest gates in the question ($0.20$--$0.26$); counts are medians over the
eight $2$-gram heads of block~2, and gates are those of one branch. The pattern
holds over $4{,}000$ questions (Figure~\ref{fig:motivation}b). Among the lookups of the
eight $2$-gram heads of block~2, rows seen more than $1{,}000$ times serve
$32\%$ but carry only $3.8\%$ of the importance, whereas rows never seen in
the counting corpus serve $14\%$ and carry $23\%$; over all $24$ heads, the
frequent rows serve $13\%$ of the lookups and carry $2.0\%$. Gate and access
count are negatively correlated in every layer, $n$-gram order, and text we
measured: Spearman $-0.40$ for the $2$-gram heads of block~2 on PopQA, and between
$-0.02$ and $-0.63$ across layers, orders, and English, Chinese, and code text
(Appendix~\ref{app:frequency}). Both factors of importance contribute: the gate
alone lowers the importance share of rows seen more than $1{,}000$ times from
$32\%$ to $11\%$, and their smaller norms lower it to $3.8\%$.

Allocating precision by frequency fails accordingly: choosing lookups by the
frequency of their rows is worse than choosing at random. Giving extra precision
to the $41\%$ of lookups with the most frequent rows, the share refined in the
1.50-bit setting of Section~\ref{sec:results-main}, removes only $26\%$ of the
error in \eqref{eq:distortion} that giving it to all lookups would remove,
whereas a random $41\%$ removes $41\%$ (Figure~\ref{fig:ranking} in
Section~\ref{sec:results-ablation}).

\paragraph{Observation 3: Importance is concentrated in a few lookups, and
context decides which.}
Importance is highly skewed across lookups. On held-out English text, the
top $10\%$ of lookups carry $48\%$ of the importance in block~2 and $69\%$ in
block~15 ($50\%$ and $47\%$ on RACE passages); on PopQA, the top $1\%$ in
block~15 alone carry $54\%$ (Appendix~\ref{app:frequency}). A small
budget could thus remove much of the error, if it reached the right lookups.
Statistics of their rows cannot find them, for two reasons. First, many of
them read rare rows, for which there is too little data to estimate an
expected gate: rows never seen in the counting corpus alone carry $23\%$ of
the importance of the $2$-gram lookups (Figure~\ref{fig:motivation}b), and
$12\%$ over all heads. Second, the gate of the same
row changes sharply with context (Figure~\ref{fig:motivation}c). In block~15,
the row that a $2$-gram head reads for \emph{a good} has a gate energy of
$45.6\times10^{-3}$ in \emph{key to a good holiday} and $0.8\times10^{-3}$ in
\emph{there is a good and clear reason}, and its lookups spread by a factor of
$4.4$ between the 10th and 90th percentile, as those of a typical row do:
across the rows looked up at least $20$ times in RACE passages, the
90th-percentile lookup of a row has $4.3$ times the gate energy of its
10th-percentile lookup in block~15, and $2.0$ times in block~2 ($8.8$ and
$2.1$ on held-out English text; Appendix~\ref{app:frequency}).

Row-level statistics of the gate are therefore no substitute for the gate
itself. Per lookup, ranking by the current gate beats ranking by the row's
expected gate in every setting we test (Section~\ref{sec:results-main}).

Together, the observations call for a scheme that reads a cheap $1$-bit code
for every lookup to obtain its gate (Observation~1), ranks lookups by that
gate rather than by statistics of their rows (Observations~2 and~3), and
spends the remaining bits only on the lookups that rank highest.
Section~\ref{sec:method} develops \method{} along these lines.

\providecommand{\method}{GateRQ}
\providecommand{\todoinline}[1]{\textcolor{red}{[TODO: #1]}}

\section{\method}
\label{sec:method}

\begin{figure}[t]
\centering
\includegraphics[width=\textwidth]{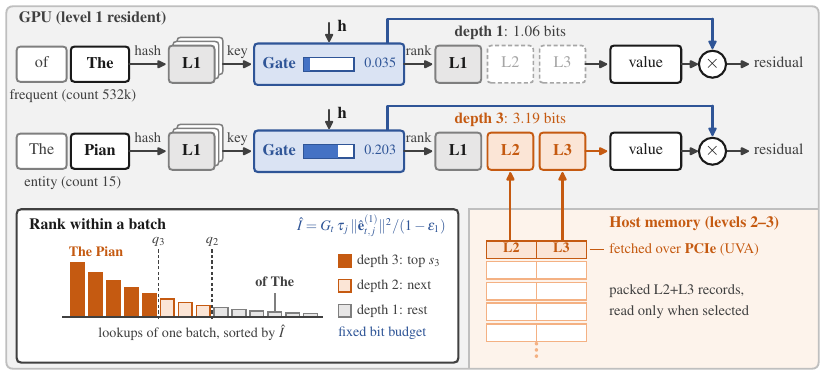}
\caption{\textbf{The gate decides how deep each lookup reads.} Every lookup
reads level~1 (L1) from GPU memory and computes its gate from it. The
importance estimate $\hat I$ of \eqref{eq:importance-estimate}, which contains
the gate, ranks the lookups of a batch: the top ones read level~3 and the next
level~2 from host memory, and the rest stop at L1, so the mean bits per lookup
stay fixed. Both lookups scale their values with the L1 gate that ranked them
(blue line). Gates and row counts are those of Figure~\ref{fig:motivation}a;
depths are illustrative; stacked L1 boxes are the rows of all hashed heads, and
the ladder and the host record are shown for one head.}
\label{fig:method-overview}
\end{figure}

\method{} has two parts (Figure~\ref{fig:method-overview}): a ladder of residual
codes, which lets a lookup's precision be raised after it has been read
(Section~\ref{sec:method-codes}), and a rule that sets each lookup's depth from
its gate (Section~\ref{sec:method-alloc}). Appendix~\ref{app:impl}
describes our implementation.

\subsection{Residual Ladder}
\label{sec:method-codes}

\method{} stores each row
$\mathbf e$ as a ladder of up to $L=3$ levels of $1$-bit residual codes
\citep{juang1982,chen2010}, each level encoding the residual left by the levels
before it:
\begin{equation}
\mathbf r_k=\mathbf e-\hat{\mathbf e}^{(k-1)},\qquad
\hat{\mathbf e}^{(k)}=\hat{\mathbf e}^{(k-1)}+a_k\mathcal Q\!\left(
\frac{\mathbf r_k}{a_k}\right),\qquad
\hat{\mathbf e}^{(0)}=\mathbf0,
\label{eq:ladder}
\end{equation}
where $\mathcal Q$ is a $1$-bit-per-element quantizer and $a_k$ is the RMS of
$\mathbf r_k$, stored as one FP16 scale per row and level. A lookup that reads
depth $\ell$, the first $\ell$ levels, reconstructs $\hat{\mathbf e}^{(\ell)}$.
Our default $\mathcal Q$ is TCQ \citep{tcq} with QTIP's trellis \citep{qtip};
including its scale, a level costs $1+16/d_e$ bits per element ($1.0625$ on
DeepSeek, $1.10$ on Qwen). The construction works with any $1$-bit quantizer;
Section~\ref{sec:results-main} substitutes integer and product codes.

Level~1 alone is a usable reconstruction, so a lookup can read it, compute its
gate (Observation~1), and only then decide how deep to go. The levels are
stored separately: level~1 stays in GPU memory, and deeper levels live in host
memory and are read only for the lookups that need them. Storage grows with
depth, to $2.13$ and $3.19$ bits per element for two and three levels, while
the reconstruction error stays close to that of direct $2$- and $3$-bit codes
(Appendix~\ref{app:ladder}).

\subsection{Per-Lookup Depth Allocation}
\label{sec:method-alloc}
\label{sec:method-formulation}

\paragraph{Objective and read rule.}
Let $\mathcal A$ be the lookups in one forward pass of a layer, and let lookup
$(t,j)$ read depth $\ell_{t,j}\in\{1,\ldots,L\}$. With $b_\ell$ the cumulative
rate through level~$\ell$ in bits per element and $\varepsilon_\ell$ the mean
relative error at that depth, \eqref{eq:distortion} gives
\begin{equation}
\min_{\{\ell_{t,j}\}}\ \sum_{(t,j)\in\mathcal A}I_{t,j}\varepsilon_{\ell_{t,j}}
\quad\text{s.t.}\quad
\frac{1}{|\mathcal A|}\sum_{(t,j)\in\mathcal A}b_{\ell_{t,j}}
\le B_{\mathrm{read}},
\label{eq:alloc}
\end{equation}
where $B_{\mathrm{read}}$ is the average read rate per lookup. With a price
$\mu\ge0$ per bit, the problem decouples over lookups, and refining a lookup
from level~$k-1$ to~$k$ is worthwhile when
\begin{equation}
I_{t,j}\Delta\varepsilon_k>\mu\Delta b_k
\quad\Longleftrightarrow\quad I_{t,j}>\mu\theta_k,
\qquad \theta_k=\frac{\Delta b_k}{\Delta\varepsilon_k},
\label{eq:read-rule}
\end{equation}
with $\Delta b_k=b_k-b_{k-1}$ and $\Delta\varepsilon_k=\varepsilon_{k-1}-\varepsilon_k$.
When each level removes less error than the previous one, as in all our
ladders, $\theta_k$ grows with $k$ and the selected sets are nested: a lookup
that reads level~$k$ also reads every level before it. This is the classical
marginal-return rule of bit allocation \citep{shoham1988}, applied to lookups
instead of table rows.

\paragraph{Importance from level~1.}
The rule needs $I_{t,j}$ before any deeper level is read, so we estimate it from
level~1:
\begin{equation}
\hat I_{t,j}=G_t\tau_j\,
\frac{\|\hat{\mathbf e}^{(1)}_{t,j}\|_2^2}{1-\varepsilon_1},
\label{eq:importance-estimate}
\end{equation}
where $G_t$ is computed from the reconstructions $\hat{\mathbf e}^{(1)}_{t,j}$ at
position~$t$ (Observation~1), $\tau_j$ is the precomputed head gain, and the last
factor estimates the row energy $\|\mathbf e_{t,j}\|_2^2$ from
$\|\hat{\mathbf e}^{(1)}_{t,j}\|_2^2$, which is about a fraction $1-\varepsilon_1$
of it.

\paragraph{Per-batch thresholds.}
A fixed price $\mu$ meets \eqref{eq:alloc} only on average over batches, so we
use per-batch quantile thresholds instead: the top $s_2$ fraction of a batch's
lookups by $\hat I$ reads at least level~2 and the top $s_3\le s_2$ reads
level~3. The fractions are fixed in advance and give every batch the same read
rate (Section~\ref{sec:setup}, Appendix~\ref{app:bits}). A lookup's depth thus
depends on the other lookups of its batch; when a batch holds a single token,
its lookups share the gate energy $G_t$ and are ranked by
$\tau_j\|\hat{\mathbf e}^{(1)}_{t,j}\|_2^2$ alone.
Algorithm~\ref{alg:forward} gives one forward pass: keys and gates come from
level~1, so deeper levels refine only the value, and injection uses the same
gates that ranked the lookups.

\begin{algorithm}[t]
\caption{One forward pass of a conditional-memory layer with \method{}.}
\label{alg:forward}
\small
\begin{algorithmic}[1]
\Require Lookups $\mathcal A$ with row IDs $r(t,j)$, branch states
$\mathbf h_t^{(m)}$, fractions $s_2,s_3$
\State Decode level~1 of every lookup to obtain $\hat{\mathbf e}^{(1)}_{t,j}$.
\State Compute keys $\mathbf k_t^{(m)}=\mathbf W_K^{(m)}\hat{\mathbf e}^{(1)}_t$, where
$\hat{\mathbf e}^{(1)}_t$ concatenates the $\hat{\mathbf e}^{(1)}_{t,j}$.
\State Compute gates $\alpha_t^{(m)}$ with the model's original gate; set
$G_t=\sum_m(\alpha_t^{(m)})^2$.
\State Compute $\hat I_{t,j}$ by \eqref{eq:importance-estimate} and the quota
thresholds $q_2\le q_3$.
\State Set $\ell_{t,j}=1+[\hat I_{t,j}\ge q_2]+[\hat I_{t,j}\ge q_3]$.
\State Read levels $2,\ldots,\ell_{t,j}$ of selected lookups; reconstruct
by \eqref{eq:ladder}.
\State Compute the value
$\mathbf v_t=\sum_j\mathbf W_V^{(j)}\hat{\mathbf e}^{(\ell_{t,j})}_{t,j}$.
\State Inject $\mathbf u_t^{(m)}=\alpha_t^{(m)}\mathbf v_t$ through the model's
original output path.
\end{algorithmic}
\end{algorithm}

\providecommand{\accfigwidth}{\textwidth}

\section{Experiments}
\label{sec:results}

\subsection{Setup}
\label{sec:setup}

\paragraph{Models and benchmarks.}
We evaluate two conditional-memory models: DeepSeek-V4.1-Flash and
Qwen3.8-Flash-Next, with row dimensions $256$ and $160$, respectively
(Section~\ref{sec:background}). Benchmarks cover factual question answering
on PopQA \citep{popqa}, TriviaQA \citep{triviaqa}, and NQ-open \citep{nq},
and Chinese multitask knowledge on CMMLU \citep{cmmlu}. All evaluations use
the full test sets, with $14{,}267$, $9{,}960$, $3{,}610$, and $11{,}582$
questions, respectively. Appendix~\ref{app:protocol} gives the protocol.

\paragraph{Baselines.}
We compare with the FP8 tables and uniform reconstruction using one, two,
or three complete residual levels. We also evaluate three selection rules
in the same setting: refine lookups by the row's expected gate energy
(\emph{prior}), by access frequency (\emph{frequency}), or uniformly at
random (\emph{random}). Each rule uses the same ladder and per-level
refinement fractions as \method{}; all mixed configurations compute keys
and injection gates from level~1, whereas uniform configurations use the
full selected depth. These are controlled selection baselines, not
separate external mixed-precision methods.

\paragraph{Implementation.}
Quantizers are calibrated on $65{,}536$ rows per head, and access counts are
collected from $1.15\times10^8$ RedPajama tokens \citep{redpajama}.
We use TCQ \citep{tcq,qtip}, INT1 with an E4M3 scale per $32$ elements, and
PQ1 with $256$ centroids per $8$-dimensional subvector and a row scale
\citep{pq}. We report logical bits per looked-up element, including scales,
and evaluate two read budgets, $1.50$ and $2.10$ bits per lookup with the TCQ
and PQ ladders. In every forward batch of a layer, a rule ranks the lookups by
its score, and fixed fractions $(s_2,s_3)$ of the lookups read level~2 and
level~3: $(41.2\%,0)$ and $(60\%,37.65\%)$ on DeepSeek, $(36.4\%,0)$ and
$(56\%,34.9\%)$ on Qwen. The budget fixes $s_2$ at $1.50$ bits; at $2.10$ bits
it leaves the split between the two deeper levels open, and we use one split
for all rules and codecs. \method{} and the three selection rules use the same
fractions and therefore read the same bits per lookup; ties, which are frequent
for access counts, are broken at random. The rows read in the runs give the
realized rate, $1.500$ and $2.100$ bits per lookup in all TCQ runs; the INT ladder,
whose levels cost more, reads more at the same fractions (Appendix~\ref{app:bits}).

\subsection{Main results}
\label{sec:results-main}

\begin{figure}[t]
\centering
\includegraphics[width=\accfigwidth]{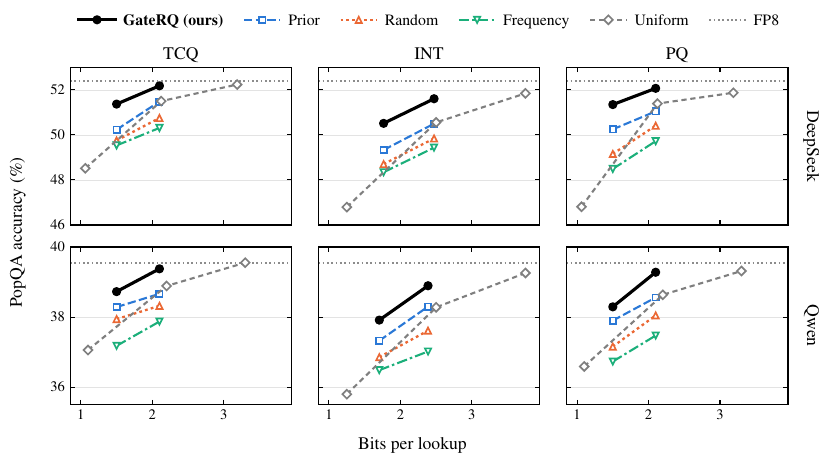}
\caption{\textbf{PopQA accuracy against bits per lookup.} Rows: DeepSeek-V4.1-Flash
and Qwen3.8-Flash-Next; columns: TCQ, INT, and PQ ladders. At each of the two read
budgets, \method{} and the three selection rules read the same bits per lookup
($1.50$ and $2.10$ with TCQ and PQ; with INT, whose levels cost $1.25$ bits each,
$1.76$ and $2.47$ on DeepSeek and $1.70$ and $2.39$ on Qwen); uniform reads one to
three full levels. Bits include scales; the dotted line is FP8 ($8.25$ bits on
DeepSeek, $8.00$ on Qwen). The y-axes are shared within each row.}
\label{fig:accuracy-popqa}
\end{figure}

Table~\ref{tab:main} gives the results with the TCQ ladder, and
Figure~\ref{fig:accuracy-popqa} shows PopQA for all three codecs (four-task means for INT
and PQ: Appendix~\ref{app:fourtask}). At the same read budget, \method{} has the best PopQA
accuracy of the four rules in all $12$ model--codec--budget settings, and all $36$ paired
comparisons with the other rules are significant, by $0.39$ to $2.86$ points
(Appendix~\ref{app:rules}). With TCQ, \method{} is within $0.2$ PopQA points of FP8 at
$2.10$ bits, which is not a significant difference, and $0.8$--$1.0$ points below it at
$1.50$ bits; its four-task mean is within $0.3$ points of FP8 at both budgets.

Reading $1.50$/$2.10$ bits instead of the full two-/three-level ladder saves
$0.63$--$1.20$ bits per element, with no statistically significant PopQA difference at
these four working points; this does not establish equivalence. On the other tasks the
differences are smaller: \method{} is significantly better than a rule in $20$ of $36$
comparisons on TriviaQA, $21$ on CMMLU, and one on NQ-open, and the only exception is
Frequency on DeepSeek TriviaQA with TCQ at $2.10$ bits (Appendix~\ref{app:rules}).
Entries above FP8 should not be read as consistent accuracy improvements.

\begin{table}[t]
\centering
\caption{\textbf{Results with the TCQ ladder.} Accuracy ($\%$) on the four benchmarks
and their mean, for the FP8 tables, uniform reconstruction with one to three levels, and
\method{} and the three selection rules at two read budgets, at which the four rules read
the same bits per lookup. \textbf{Read}: bits used per looked-up element. Bold: best among
the four rules. PopQA half-widths of the paired $95\%$ intervals against FP8:
$0.28$--$0.47$ points. Paired differences between the rules: Appendix~\ref{app:rules}.}
\label{tab:main}
\footnotesize
\setlength{\tabcolsep}{4pt}
\renewcommand{\arraystretch}{0.92}
\begin{tabular}{lrrrrrr}
\toprule
Method & Read & PopQA & TriviaQA & CMMLU & NQ-open & Avg. \\
\midrule
\multicolumn{7}{l}{\textit{DeepSeek-V4.1-Flash}} \\
FP8 & 8.25 & 52.39 & 82.78 & 89.48 & 32.71 & 64.34 \\
\midrule
Uniform, level~1 & 1.06 & 48.52 & 82.92 & 88.83 & 32.96 & 63.31 \\
Uniform, levels~1--2 & 2.13 & 51.50 & 82.97 & 89.26 & 32.74 & 64.12 \\
Uniform, levels~1--3 & 3.19 & 52.24 & 82.87 & 89.52 & 33.13 & 64.44 \\
\cmidrule(lr){1-7}
\multicolumn{7}{l}{\textit{$1.50$ bits per lookup}} \\
\ours Gate (\method{}) & 1.50 & \textbf{51.37} & 82.97 & \textbf{89.36} & \textbf{33.02} & \textbf{64.18} \\
Prior & 1.50 & 50.23 & 82.88 & 89.24 & 32.99 & 63.84 \\
Random & 1.50 & 49.75 & 83.02 & 89.11 & 32.83 & 63.68 \\
Frequency & 1.50 & 49.53 & \textbf{83.12} & 89.04 & 32.99 & 63.67 \\
\cmidrule(lr){1-7}
\multicolumn{7}{l}{\textit{$2.10$ bits per lookup}} \\
\ours Gate (\method{}) & 2.10 & \textbf{52.18} & 82.88 & 89.43 & \textbf{33.35} & \textbf{64.46} \\
Prior & 2.10 & 51.48 & 82.90 & \textbf{89.52} & 32.99 & 64.22 \\
Random & 2.10 & 50.75 & 83.01 & 89.30 & 32.99 & 64.02 \\
Frequency & 2.10 & 50.31 & \textbf{83.23} & 89.06 & 33.13 & 63.93 \\
\midrule
\multicolumn{7}{l}{\textit{Qwen3.8-Flash-Next}} \\
FP8 & 8.00 & 39.55 & 75.44 & 88.80 & 32.94 & 59.18 \\
\midrule
Uniform, level~1 & 1.10 & 37.06 & 74.40 & 88.16 & 33.02 & 58.16 \\
Uniform, levels~1--2 & 2.20 & 38.89 & 75.37 & 88.62 & 33.19 & 59.02 \\
Uniform, levels~1--3 & 3.30 & 39.55 & 75.52 & 88.68 & 33.38 & 59.28 \\
\cmidrule(lr){1-7}
\multicolumn{7}{l}{\textit{$1.50$ bits per lookup}} \\
\ours Gate (\method{}) & 1.50 & \textbf{38.73} & \textbf{75.55} & \textbf{88.59} & 32.80 & \textbf{58.91} \\
Prior & 1.50 & 38.28 & 75.26 & 88.51 & 32.85 & 58.73 \\
Random & 1.50 & 37.94 & 74.85 & 88.42 & 32.77 & 58.50 \\
Frequency & 1.50 & 37.18 & 74.69 & 88.19 & \textbf{33.30} & 58.34 \\
\cmidrule(lr){1-7}
\multicolumn{7}{l}{\textit{$2.10$ bits per lookup}} \\
\ours Gate (\method{}) & 2.10 & \textbf{39.38} & \textbf{75.68} & \textbf{88.64} & \textbf{33.21} & \textbf{59.23} \\
Prior & 2.10 & 38.66 & 75.39 & 88.54 & 33.19 & 58.95 \\
Random & 2.10 & 38.32 & 75.18 & 88.52 & 33.07 & 58.77 \\
Frequency & 2.10 & 37.87 & 74.91 & 88.26 & 32.49 & 58.38 \\
\bottomrule
\end{tabular}

\end{table}

\subsection{Ablation}
\label{sec:results-ablation}

\noindent\begin{minipage}[t]{0.51\textwidth}
\vspace{0pt}
\paragraph{Which lookups to refine.}
Figure~\ref{fig:ranking} compares four rules for choosing the lookups that read
level~2: the estimated importance $\hat I$ of \eqref{eq:importance-estimate}
(Gate), the same score with the gate replaced by the expected gate of the row
(Prior), a random choice, and the access count of the row (Frequency). Under
the error proxy of \eqref{eq:distortion}, ranking by $\hat I$ removes $93\%$ of
the error that refining every lookup would remove at the $1.50$-bit setting,
where $41.2\%$ of the lookups read level~2, against $70\%$ for Prior, $41\%$
for Random, and $26\%$ for Frequency. The measured PopQA accuracy of the TCQ
runs gives the same order: the four rules recover $96\%$, $58\%$, $41\%$, and
$34\%$ of the gap between level~1 and two full levels. The proxy lies inside the
paired $95\%$ interval of the measurement for Gate, Random, and Frequency, and
overestimates Prior ($58\%$, interval $47$--$68\%$).
\end{minipage}\hfill
\begin{minipage}[t]{0.45\textwidth}
\vspace{0pt}
\centering
\includegraphics[width=\linewidth]{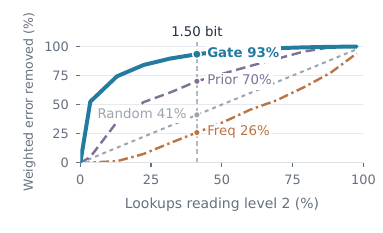}
\captionsetup{type=figure,font=small,justification=raggedright}
\captionof{figure}{\textbf{Gate ranking removes the most error.} Weighted error of
\eqref{eq:distortion} removed, relative to refining every lookup to level~2,
against the share of lookups reading level~2 (offline proxy; PopQA, DeepSeek).
Dashed line: the $1.50$-bit setting.}
\label{fig:ranking}
\end{minipage}

\subsection{Efficiency}
\label{sec:results-efficiency}

\begin{figure}[t]
\centering
\includegraphics[width=\textwidth]{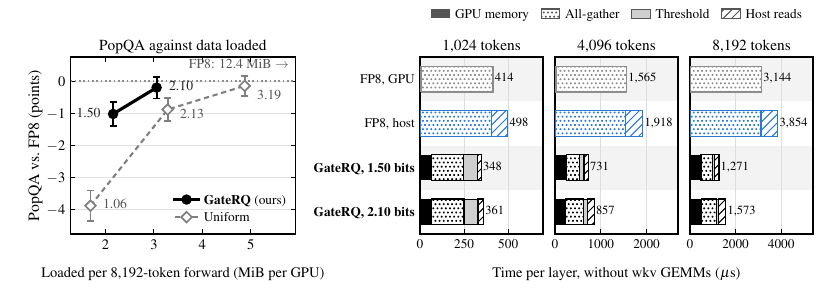}
\caption{\textbf{What a forward pass loads and what the table side costs}
(DeepSeek-V4.1-Flash, TP8). \emph{Left:} PopQA change from FP8
(paired $95\%$ intervals; the TCQ runs of
Table~\ref{tab:main}) against the table data one GPU loads in
a forward pass of $8{,}192$ tokens, which is proportional to the bits per lookup
(labels). Uniform reads one to three full levels; points are offset horizontally
for legibility. \emph{Right:} time of the table side of one Engram layer in a
pass of $1{,}024$, $4{,}096$ and $8{,}192$ tokens, per rank, split by where it
goes. FP8 tables are in GPU memory or in pinned host memory; \method{} reads the
deeper records of each rank's own heads from host memory. The wkv projection
($7.4$~ms at $8{,}192$ tokens; $-6\%$ to $+3\%$ when split as in \method{}) is
not included.}
\label{fig:load-time}
\end{figure}

Figure~\ref{fig:load-time} shows what a forward pass loads and what the table
side of an Engram layer costs. With \method{}, a forward pass of $8{,}192$
tokens loads $2.25$--$3.15$~MiB of table data per GPU instead of $12.4$~MiB,
and the table side takes $1.3$--$1.6$~ms instead of $3.1$~ms with FP8 tables
in GPU memory and $3.9$~ms in host memory. The gap shrinks with fewer tokens:
$0.35$--$0.36$~ms against $0.41$--$0.50$~ms at $1{,}024$ tokens. These timings
read the deeper records of each rank's own heads; the serving runs behind
Table~\ref{tab:efficiency} (Appendix~\ref{app:serving}) use a simpler variant in
which every rank reads the deeper records of all heads ($2.9$--$4.6$~ms,
identical outputs). In either case the table side is about $1\%$ of a prefill
step, so these differences do not show up in end-to-end throughput.
Implementation details are given in Appendix~\ref{app:impl}.

\section{Conclusion}
\label{sec:conclusion}

Quantizing conditional-memory tables comes down to deciding which lookups deserve more
bits, and the answer is the gate, not access frequency: the gate scales each retrieved row
and can be computed from a $1$-bit code. \method{} stores each row as a ladder of $1$-bit
residual codes and reads deeper levels only for the lookups that the gate ranks highest.
Across DeepSeek-V4.1-Flash and Qwen3.8-Flash-Next and three codecs, it beats the other
selection rules in all $36$ paired PopQA comparisons at the same bits read per lookup, and
with TCQ it is within $0.2$ PopQA points of FP8 at $2.10$ bits, while cutting DeepSeek's
table memory from $23.6$ to $3.0$~GiB per GPU and raising vLLM throughput by
$11\%$--$31\%$ over FP8 tables in GPU memory.

\FloatBarrier
\bibliographystyle{unsrtnat}
\bibliography{refs}

\begin{thebibliography}{27}
\providecommand{\natexlab}[1]{#1}
\providecommand{\url}[1]{\texttt{#1}}
\expandafter\ifx\csname urlstyle\endcsname\relax
  \providecommand{\doi}[1]{doi: #1}\else
  \providecommand{\doi}{doi: \begingroup \urlstyle{rm}\Url}\fi

\bibitem[Cheng et~al.(2026)Cheng, Tian, Zeng, Dai, Chen, Wang, Xie, Huang, Yu,
  Deng, Zhou, Zhao, Hao, Li, Zhang, Zhang, Wei, Xu, Zhang, Zhao, and
  Liang]{engram}
Xin Cheng, Rui Tian, Wangding Zeng, Damai Dai, Qinyu Chen, Bingxuan Wang,
  Zhenda Xie, Kezhao Huang, Xingkai Yu, Chengqi Deng, Shangyan Zhou, Chenggang
  Zhao, Zhewen Hao, Yukun Li, Han Zhang, Zhengyan Zhang, Yixuan Wei, M.~Y. Xu,
  Huishuai Zhang, Dongyan Zhao, and Wenfeng Liang.
\newblock Conditional memory via scalable lookup: A new axis of sparsity for
  large language models.
\newblock \emph{arXiv preprint arXiv:2601.07372}, 2026.

\bibitem[{Qwen Team}(2026)]{qwen38next}
{Qwen Team}.
\newblock On the design of {Qwen3.8-Next} architecture: Evaluation, efficiency,
  and training stability.
\newblock \emph{arXiv preprint arXiv:2608.30320}, 2026.

\bibitem[Mallen et~al.(2023)Mallen, Asai, Zhong, Das, Khashabi, and
  Hajishirzi]{popqa}
Alex Mallen, Akari Asai, Victor Zhong, Rajarshi Das, Daniel Khashabi, and
  Hannaneh Hajishirzi.
\newblock When not to trust language models: Investigating effectiveness of
  parametric and non-parametric memories.
\newblock In \emph{Proceedings of the 61st Annual Meeting of the Association
  for Computational Linguistics}, 2023.

\bibitem[Ginart et~al.(2021)Ginart, Naumov, Mudigere, Yang, and
  Zou]{ginart2021mixed}
Antonio~A. Ginart, Maxim Naumov, Dheevatsa Mudigere, Jiyan Yang, and James Zou.
\newblock Mixed dimension embeddings with application to memory-efficient
  recommendation systems.
\newblock In \emph{IEEE International Symposium on Information Theory
  ({ISIT})}, 2021.

\bibitem[Yang et~al.(2020)Yang, Huang, Park, Tang, and Tulloch]{yang2020mixed}
Jie~Amy Yang, Jianyu Huang, Jongsoo Park, Ping Tak~Peter Tang, and Andrew
  Tulloch.
\newblock Mixed-precision embedding using a cache.
\newblock \emph{arXiv preprint arXiv:2010.11305}, 2020.

\bibitem[Li et~al.(2024{\natexlab{a}})Li, Guo, Tang, Tang, Hou, Li, and
  Zhang]{li2024embedding}
Shiwei Li, Huifeng Guo, Xing Tang, Ruiming Tang, Lu~Hou, Ruixuan Li, and Rui
  Zhang.
\newblock Embedding compression in recommender systems: A survey.
\newblock \emph{{ACM} Computing Surveys}, 56\penalty0 (5):\penalty0
  130:1--130:21, 2024{\natexlab{a}}.

\bibitem[Marcellin and Fischer(1990)]{tcq}
Michael~W. Marcellin and Thomas~R. Fischer.
\newblock Trellis coded quantization of memoryless and {Gauss-Markov} sources.
\newblock \emph{IEEE Transactions on Communications}, 38\penalty0 (1):\penalty0
  82--93, 1990.

\bibitem[Tseng et~al.(2024)Tseng, Sun, Hou, and Sa]{qtip}
Albert Tseng, Qingyao Sun, David Hou, and Christopher~De Sa.
\newblock {QTIP}: Quantization with trellises and incoherence processing.
\newblock In \emph{Advances in Neural Information Processing Systems}, 2024.

\bibitem[Kwon et~al.(2023)Kwon, Li, Zhuang, Sheng, Zheng, Yu, Gonzalez, Zhang,
  and Stoica]{pagedattention}
Woosuk Kwon, Zhuohan Li, Siyuan Zhuang, Ying Sheng, Lianmin Zheng, Cody~Hao Yu,
  Joseph Gonzalez, Hao Zhang, and Ion Stoica.
\newblock Efficient memory management for large language model serving with
  pagedattention.
\newblock In \emph{Proceedings of the 29th Symposium on Operating Systems
  Principles, {SOSP} 2023, Koblenz, Germany, October 23-26, 2023}, pages
  611--626. {ACM}, 2023.
\newblock \doi{10.1145/3600006.3613165}.

\bibitem[Xie et~al.(2025)Xie, Wei, Cao, Zhao, Deng, Li, Dai, Gao, Chang, Yu,
  Zhao, Zhou, Xu, Zhang, Zeng, Hu, Wang, Yuan, Wang, and Liang]{xie2025mhc}
Zhenda Xie, Yixuan Wei, Huanqi Cao, Chenggang Zhao, Chengqi Deng, Jiashi Li,
  Damai Dai, Huazuo Gao, Jiang Chang, Kuai Yu, Liang Zhao, Shangyan Zhou, Zhean
  Xu, Zhengyan Zhang, Wangding Zeng, Shengding Hu, Yuqing Wang, Jingyang Yuan,
  Lean Wang, and Wenfeng Liang.
\newblock {mHC}: Manifold-constrained hyper-connections.
\newblock \emph{arXiv preprint arXiv:2512.24880}, 2025.

\bibitem[Juang and {Gray, Jr.}(1982)]{juang1982}
Biing-Hwang Juang and Augustine~H. {Gray, Jr.}
\newblock Multiple stage vector quantization for speech coding.
\newblock In \emph{IEEE International Conference on Acoustics, Speech, and
  Signal Processing ({ICASSP})}, pages 597--600, 1982.

\bibitem[Chen et~al.(2010)Chen, Guan, and Wang]{chen2010}
Yongjian Chen, Tao Guan, and Cheng Wang.
\newblock Approximate nearest neighbor search by residual vector quantization.
\newblock \emph{Sensors}, 10\penalty0 (12):\penalty0 11259--11273, 2010.

\bibitem[Shoham and Gersho(1988)]{shoham1988}
Yair Shoham and Allen Gersho.
\newblock Efficient bit allocation for an arbitrary set of quantizers.
\newblock \emph{IEEE Transactions on Acoustics, Speech, and Signal Processing},
  36\penalty0 (9):\penalty0 1445--1453, 1988.

\bibitem[Joshi et~al.(2017)Joshi, Choi, Weld, and Zettlemoyer]{triviaqa}
Mandar Joshi, Eunsol Choi, Daniel~S. Weld, and Luke Zettlemoyer.
\newblock {TriviaQA}: A large scale distantly supervised challenge dataset for
  reading comprehension.
\newblock In \emph{Proceedings of the 55th Annual Meeting of the Association
  for Computational Linguistics}, 2017.

\bibitem[Kwiatkowski et~al.(2019)Kwiatkowski, Palomaki, Redfield, Collins,
  Parikh, Alberti, Epstein, Polosukhin, Devlin, Lee, Toutanova, Jones, Kelcey,
  Chang, Dai, Uszkoreit, Le, and Petrov]{nq}
Tom Kwiatkowski, Jennimaria Palomaki, Olivia Redfield, Michael Collins, Ankur
  Parikh, Chris Alberti, Danielle Epstein, Illia Polosukhin, Jacob Devlin,
  Kenton Lee, Kristina Toutanova, Llion Jones, Matthew Kelcey, Ming-Wei Chang,
  Andrew~M. Dai, Jakob Uszkoreit, Quoc Le, and Slav Petrov.
\newblock Natural questions: A benchmark for question answering research.
\newblock \emph{Transactions of the Association for Computational Linguistics},
  7:\penalty0 452--466, 2019.

\bibitem[Li et~al.(2024{\natexlab{b}})Li, Zhang, Koto, Yang, Zhao, Gong, Duan,
  and Baldwin]{cmmlu}
Haonan Li, Yixuan Zhang, Fajri Koto, Yifei Yang, Hai Zhao, Yeyun Gong, Nan
  Duan, and Timothy Baldwin.
\newblock {CMMLU}: Measuring massive multitask language understanding in
  {Chinese}.
\newblock In \emph{Findings of the Association for Computational Linguistics:
  {ACL} 2024}, 2024{\natexlab{b}}.

\bibitem[{Together Computer}(2023)]{redpajama}
{Together Computer}.
\newblock {RedPajama}: An open source recipe to reproduce {LLaMA} training
  dataset.
\newblock \url{https://github.com/togethercomputer/RedPajama-Data}, April 2023.

\bibitem[J{\'e}gou et~al.(2011)J{\'e}gou, Douze, and Schmid]{pq}
Herv{\'e} J{\'e}gou, Matthijs Douze, and Cordelia Schmid.
\newblock Product quantization for nearest neighbor search.
\newblock \emph{IEEE Transactions on Pattern Analysis and Machine
  Intelligence}, 33\penalty0 (1):\penalty0 117--128, 2011.

\bibitem[Zhou et~al.(2026)Zhou, Gu, Iacovides, Qiu, Zhao, and
  Mandic]{zhou2026tngram}
Wuyang Zhou, Yuxuan Gu, Giorgos Iacovides, Yuning Qiu, Qibin Zhao, and Danilo
  Mandic.
\newblock Tensorizing {Engram}: Sharing latents across {N}-gram embeddings is
  beneficial in {LLMs}.
\newblock \emph{arXiv preprint arXiv:2606.08347}, 2026.

\bibitem[Yang et~al.(2026)Yang, Zhou, Miao, and Wu]{yang2026factorengram}
Bowen Yang, Jingbo Zhou, Qinghong Miao, and Hua Wu.
\newblock {FactorEngram}: Factorized {N}-gram memory with basis-level gating
  for language models.
\newblock \emph{arXiv preprint arXiv:2609.35578}, 2026.

\bibitem[Zheng et~al.(2026)Zheng, Xia, Wang, and Ren]{zheng2026lngram}
Yunao Zheng, Guoyang Xia, Xiaojie Wang, and Lei Ren.
\newblock {Lngram}: {N}-gram conditional memory in latent space.
\newblock \emph{arXiv preprint arXiv:2605.24869}, 2026.

\bibitem[Lin(2026)]{lin2026hottier}
Tao Lin.
\newblock A collision-free hot-tier extension for {Engram}-style conditional
  memory: A controlled study of training dynamics.
\newblock \emph{arXiv preprint arXiv:2601.16531}, 2026.

\bibitem[Ma et~al.(2026{\natexlab{a}})Ma, Huang, Jiang, and
  Shao]{ma2026tfengram}
Yutang Ma, Kecheng Huang, Xikun Jiang, and Zili Shao.
\newblock {TF-Engram}: A train-free {Engram} with {SSD}-backed memory for large
  language models.
\newblock \emph{arXiv preprint arXiv:2607.07388}, 2026{\natexlab{a}}.

\bibitem[Ma et~al.(2026{\natexlab{b}})Ma, Ma, Su, Zha, Zhao, Shang, Yi, Liu,
  Cao, Wu, Dou, Liu, Kuang, and Luo]{ma2026cxl}
Ruiyang Ma, Teng Ma, Zhiyuan Su, Hantian Zha, Xinpeng Zhao, Xuchun Shang,
  Xingrui Yi, Zheng Liu, Zhu Cao, An~Wu, Zhichong Dou, Ziqian Liu, Daikang
  Kuang, and Guojie Luo.
\newblock Pooling {Engram} conditional memory in large language models using
  {CXL}.
\newblock In \emph{Proceedings of the 6th Workshop on Machine Learning and
  Systems}, 2026{\natexlab{b}}.

\bibitem[Bi et~al.(2026)Bi, Chen, Wang, Peng, Hong, Wang, Chu, Zhu, and
  Lou]{bi2026memory}
Zhen Bi, Xueshu Chen, Yan Wang, Zhizhi Peng, Haosen Hong, Zhen Wang, Zhixuan
  Chu, Bingyu Zhu, and Jungang Lou.
\newblock Memory is not always needed: Characterizing conditional memory in
  scientific reasoning.
\newblock \emph{arXiv preprint arXiv:2608.23982}, 2026.

\bibitem[Equitz and Cover(1991)]{equitz1991}
William H.~R. Equitz and Thomas~M. Cover.
\newblock Successive refinement of information.
\newblock \emph{IEEE Transactions on Information Theory}, 37\penalty0
  (2):\penalty0 269--275, 1991.

\bibitem[Gao et~al.(2024)Gao, Tow, Abbasi, Biderman, Black, DiPofi, Foster,
  Golding, Hsu, Noac'h, Li, McDonell, Muennighoff, Ociepa, Phang, Reynolds,
  Schoelkopf, Skowron, Sutawika, Tang, Thite, Wang, Wang, and Zou]{lmeval}
Leo Gao, Jonathan Tow, Baber Abbasi, Stella Biderman, Sid Black, Anthony
  DiPofi, Charles Foster, Laurence Golding, Jeffrey Hsu, Alain~Le Noac'h,
  Haonan Li, Kyle McDonell, Niklas Muennighoff, Chris Ociepa, Jason Phang,
  Laria Reynolds, Hailey Schoelkopf, Aviya Skowron, Lintang Sutawika, Eric
  Tang, Anish Thite, Ben Wang, Kevin Wang, and Andy Zou.
\newblock A framework for few-shot language model evaluation.
\newblock Zenodo, 2024.

\end{thebibliography}

\clearpage
\appendix
\providecommand{\method}{GateRQ}

\section{Related Work}
\label{app:related}

\paragraph{Efficient Engram.}
Engram \citep{engram} keeps its tables off the backbone's critical path: row
IDs depend only on tokens, so the tables can sit in host memory and be
prefetched. Follow-up work revisits how the tables are parameterized, stored,
and used.

\emph{Different parameterizations, trained with the model.} TN-gram replaces
the hash tables with CP factors shared across $n$-gram orders and matches
Engram's validation loss with 20--27\% fewer memory parameters
\citep{zhou2026tngram}. FactorEngram stores sparse coefficients over a shared
dictionary and gates each dictionary vector separately
\citep{yang2026factorengram}, and Lngram addresses its tables with discrete
codes learned from hidden states instead of token IDs
\citep{zheng2026lngram}; both improve quality at the same parameter budget.
Engram-Nine gives the most frequent $n$-grams collision-free rows and finds no
improvement in loss at an equal parameter count \citep{lin2026hottier}, in
line with our observation that frequency is a poor guide to where capacity
matters. In that small model, trained on 82M tokens, the gate favors frequent
$n$-grams, a preference that shrinks during training; in the production
models we study, the relation is reversed (Section~\ref{sec:observations}).
All of these designs require training, and the largest model among them has
8B parameters.

\emph{Cheaper storage.} TF-Engram builds phrase memories offline without
training and serves them from a GPU--DRAM--SSD hierarchy, hiding SSD latency
with predictive prefetching driven by an early-exit head
\citep{ma2026tfengram}. \citet{ma2026cxl} pool Engram tables in CXL memory
shared by several hosts and reach near-DRAM end-to-end performance in SGLang.

\emph{Selective use.} \citet{bi2026memory} find that conditional memory helps
some inputs and hurts others in scientific reasoning, and route whole inputs
and layers to or away from memory with task-specific input features.

\method{} is complementary to these lines. It compresses the tables of an
existing checkpoint after training, without gradient updates, and sets the
precision of each lookup from the model's own gate. It reduces what a storage
hierarchy must hold and transfer, it makes a finer-grained decision than
input-level routing, and it could in principle be extended to the trained
designs above that keep large tables and context gates.

\providecommand{\method}{GateRQ}

\section{Models and Tables}
\label{app:models}

DeepSeek-V4.1-Flash holds $768{,}022{,}850$ rows in two Engram layers
(Table~\ref{tab:models}). In FP8, with an E8M0 scale per $32$ elements, they take
$202.8$~GB ($188.8$~GiB), $39.7\%$ of the $510.3$~GB checkpoint.
Qwen3.8-Flash-Next holds $320{,}001{,}536$ rows in one PLE layer, $51.2$~GB in FP8
and $27.6\%$ of its $185.5$~GB checkpoint.

\begin{table}[h]
\centering
\caption{\textbf{The conditional-memory layers of the two models.}}
\label{tab:models}
\small
\begin{tabular}{lcc}
\toprule
 & DeepSeek-V4.1-Flash & Qwen3.8-Flash-Next \\
\midrule
Memory layers & Engram, blocks 2 and 15 & PLE, block 2 \\
Hash heads per layer & 24 (eight each of 2-, 3-, 4-grams) & 16 (eight each of 2-, 3-grams) \\
Row dimension $d_e$ & 256 & 160 \\
Rows, all layers & $768{,}022{,}850$ & $320{,}001{,}536$ \\
FP8 table & $202.8$ GB, $8.25$ bits per element & $51.2$ GB, $8.00$ bits per element \\
Share of the checkpoint & $39.7\%$ of $510.3$ GB & $27.6\%$ of $185.5$ GB \\
Branches $M$ & 4 & 4 \\
\bottomrule
\end{tabular}
\end{table}

\paragraph{Zeroing the tables.}
Table~\ref{tab:zero} shows how much of the models' accuracy the tables carry.
Setting every row to zero costs DeepSeek $36.76$ PopQA points, about $70\%$ of its
accuracy, and Qwen $12.90$ points, about $33\%$; on the other three benchmarks it
costs $8.1$--$14.4$ points and $4.5$--$9.7$ points, respectively.

\begin{table}[h]
\centering
\caption{\textbf{Zeroing the tables.} Change from FP8 in points when every row of
the tables is set to zero; PopQA with the paired $95\%$ interval over questions.}
\label{tab:zero}
\small
\setlength{\tabcolsep}{5pt}
\begin{tabular}{lcccc}
\toprule
 & PopQA & TriviaQA & CMMLU & NQ-open \\
\midrule
DeepSeek-V4.1-Flash & $-36.76$ $[-37.60, -35.96]$ & $-14.37$ & $-8.12$ & $-9.34$ \\
Qwen3.8-Flash-Next & $-12.90$ $[-13.58, -12.22]$ & $-9.74$ & $-4.51$ & $-5.18$ \\
\bottomrule
\end{tabular}
\end{table}

\section{Supporting Analyses for Section~\ref{sec:motivation}}
\label{app:analysis}

This appendix gives the measurements behind the observations of
Section~\ref{sec:motivation}. Its records come from the FP8 models: for every
lookup they hold the row, the gate of its position, and the access count of the row
in the counting corpus ($1.15\times10^8$ RedPajama tokens). Five sets of positions
are used: held-out English documents (about $0.7$ million positions),
Chinese ($0.12$ million) and code ($0.08$ million) documents, RACE passages
($677{,}076$ positions in $1{,}994$ passages), and the prompts of $4{,}000$ PopQA
questions ($1{,}279{,}844$ positions; a prompt holds the question and its $15$
few-shot examples).

\subsection{Loss by Subject Popularity}
\label{app:popularity}

\begin{figure}[h]
\centering
\includegraphics[width=0.92\textwidth]{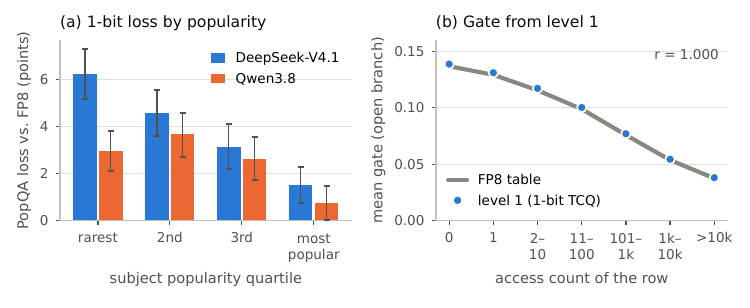}
\caption{\textbf{Loss by popularity, and the gate from level~1.}
(a)~PopQA accuracy lost when only level~1 (1-bit TCQ) is read, by quartile of
subject popularity, with paired $95\%$ intervals.
(b)~DeepSeek, block~2, $2$-gram heads, held-out English text: mean gate of the
open branch by access count of the row, computed from the FP8 table (line) and from
level~1 (dots); $r$ is the correlation over positions
(Appendix~\ref{app:gate}).}
\label{fig:app-motivation}
\end{figure}

PopQA gives the popularity of each question's subject, the monthly page views of its
Wikipedia page. Table~\ref{tab:popularity} splits the $14{,}267$ questions into four
quartiles by this number and shows the accuracy of FP8 and the change when only
level~1 is used (Figure~\ref{fig:app-motivation}a). On DeepSeek the loss falls
steadily from $6.24$ points on the rarest quartile to $1.51$ on the most popular.
On Qwen it does not: the loss is largest on the second quartile ($3.67$ points),
which also has the lowest FP8 accuracy ($26.85\%$), and exceeds the loss on the
rarest quartile ($2.95$), so the loss is spread over the three less popular
quartiles. In both models the most popular quartile loses least.

\begin{table}[h]
\centering
\caption{\textbf{PopQA with level~1 only, by popularity of the subject.} FP8
accuracy and the change when only level~1 is read, with the paired $95\%$ interval
over questions (quartiles of the subject's monthly Wikipedia page views).}
\label{tab:popularity}
\small
\setlength{\tabcolsep}{3pt}
\begin{tabular}{llrrrcrrc}
\toprule
 & & & \multicolumn{3}{c}{DeepSeek-V4.1-Flash} & \multicolumn{3}{c}{Qwen3.8-Flash-Next} \\
\cmidrule(lr){4-6}\cmidrule(lr){7-9}
Quartile & Page views & Questions & FP8 & Level 1 & 95\% interval & FP8 & Level 1 & 95\% interval \\
\midrule
Rarest & 2--235 & 3{,}559 & 42.23 & $-6.24$ & $[-7.31, -5.17]$ & 29.53 & $-2.95$ & $[-3.79, -2.11]$ \\
Second & 235--975 & 3{,}572 & 44.15 & $-4.59$ & $[-5.57, -3.58]$ & 26.85 & $-3.67$ & $[-4.59, -2.72]$ \\
Third & 975--5{,}784 & 3{,}569 & 51.67 & $-3.14$ & $[-4.09, -2.19]$ & 36.34 & $-2.63$ & $[-3.56, -1.71]$ \\
Most popular & $>5{,}784$ & 3{,}567 & 71.49 & $-1.51$ & $[-2.27, -0.76]$ & 65.49 & $-0.73$ & $[-1.46, 0.00]$ \\
\bottomrule
\end{tabular}

\end{table}

\subsection{Assumptions of \texorpdfstring{\eqref{eq:distortion}}{Equation (5)}}
\label{app:assumptions}

\eqref{eq:distortion} treats the error of a row as isotropic after the value
projection, so that it contributes $\tau_j\|\boldsymbol\delta\|_2^2$ to the
value, and the errors of different lookups as uncorrelated. We test the first
assumption on the checkpoints; the second is tested end to end by the predictions
of Appendix~\ref{app:predictions}.

\paragraph{Rows.}
The rows of the DeepSeek tables are close to independent Gaussian vectors: the ratio
of a row's largest absolute value to its RMS has median $3.005$, the value for a
$256$-dimensional Gaussian, and the standard deviations of the dimensions differ
by at most $5\%$. The Qwen tables behave alike.

\paragraph{Value projection.}
Table~\ref{tab:gram} gives statistics of the Gram matrix
$\mathbf W_V^{(j)\top}\mathbf W_V^{(j)}$, averaged over the eight heads of each
order. Its effective rank is $230$--$252$ out of $256$ on DeepSeek and $154$ out of
$160$ on Qwen, so errors reach the value nearly isotropically, and weighting a row
error by the projection raises it by only about $1\%$. The gain $\tau_j$ falls with
the order: the same row error matters $2.3$ times (block~2) and $2.9$ times
(block~15) more in a $2$-gram head than in a $4$-gram head, which is why the
importance carries $\tau_j$.

\paragraph{Key projection.}
Weighted by the channel weights of the gate, the key part of the projection is
almost rank one (effective rank $1.2$--$2.4$ on DeepSeek, $52.5$ on Qwen; the ten
leading directions hold $98$--$100\%$ of its energy on DeepSeek). The gate thus
reads a few directions of the row, and the key-weighted error of the quantized rows
is only $0.17$--$0.26$ times their unweighted error, which helps to explain why the
gate error is small (Observation~1).

\begin{table}[h]
\centering
\caption{\textbf{Value and key projections of DeepSeek's heads.} Means over the
eight heads of each order. $\tau$: gain of the head, the trace of
$\mathbf W_V^\top\mathbf W_V$ over $d_e$. Effective rank of the value Gram matrix
(out of $256$) and of the gate-weighted key Gram matrix; energy of the key Gram
matrix in its ten leading directions.}
\label{tab:gram}
\small
\begin{tabular}{llrrrr}
\toprule
Block & Order & $\tau$ & Value rank & Key rank & Key top-10 energy (\%) \\
\midrule
\multirow{3}{*}{2} & 2 & 2.87 & 244 & 1.56 & 99.7 \\
 & 3 & 1.48 & 243 & 1.24 & 99.7 \\
 & 4 & 1.25 & 242 & 1.24 & 99.9 \\
\multirow{3}{*}{15} & 2 & 3.60 & 252 & 2.35 & 98.5 \\
 & 3 & 1.85 & 241 & 1.36 & 98.2 \\
 & 4 & 1.26 & 230 & 1.25 & 97.8 \\
\bottomrule
\end{tabular}

\end{table}

\subsection{Gates Computed from Level~1}
\label{app:gate}

Observation~1 compares the gates computed from level~1 with those computed from the
FP8 table, on records of held-out text, and measures the accuracy that follows.

\paragraph{Gate values.}
Table~\ref{tab:gate-fidelity} gives the correlation over positions. On DeepSeek
it is $0.9996$--$0.9998$ in block~2 and $0.989$--$0.990$ in block~15; the level-1
gate energy is $2$--$5\%$ above the FP8 one on average. On Qwen the correlation of
the gate energy is $0.990$--$0.995$. A scale on the key weights, fitted to align
the level-1 gate of the open branch with the FP8 gate, is between $1.00$ and $1.03$
in all six DeepSeek cases, so there is no systematic bias to calibrate away.

\begin{table}[h]
\centering
\caption{\textbf{Gates from level~1 against gates from FP8.} DeepSeek: correlation
over positions of the gate of the open branch, and ratio of its mean squared gate
(level~1 over FP8). Qwen: correlation of the gate energy, and overlap of the top
$28\%$ of positions by gate energy under the two gates.}
\label{tab:gate-fidelity}
\small
\setlength{\tabcolsep}{5pt}
\begin{tabular}{lcccccc}
\toprule
 & \multicolumn{2}{c}{DeepSeek, block 2} & \multicolumn{2}{c}{DeepSeek, block 15} & \multicolumn{2}{c}{Qwen} \\
\cmidrule(lr){2-3}\cmidrule(lr){4-5}\cmidrule(lr){6-7}
Text & Correlation & Ratio & Correlation & Ratio & Correlation & Overlap (\%) \\
\midrule
English & 0.9998 & 1.025 & 0.9893 & 1.017 & 0.9896 & 95.9 \\
Chinese & 0.9996 & 1.026 & 0.9899 & 1.029 & 0.9925 & 95.9 \\
Code & 0.9998 & 1.022 & 0.9901 & 1.049 & 0.9947 & 98.0 \\
\bottomrule
\end{tabular}

\end{table}

\paragraph{Selected lookups.}
What matters for allocation is which lookups the two gates select. On DeepSeek, we
rank the lookups of all $24$ heads by the importance of \eqref{eq:distortion}, once with the
level-1 gate and once with the FP8 gate, on held-out English text, and compare the
sets that receive a deeper read (Table~\ref{tab:overlap}). The lookups that the
level-1 gate selects carry $99.2\%$ (block~15, top $10\%$) to $99.99\%$ (block~2, top
$41.2\%$) of the importance carried by those that the FP8 gate selects. The overlaps
were computed offline on the English records of Table~\ref{tab:gate-fidelity}.

\begin{table}[h]
\centering
\caption{\textbf{Agreement of the lookups selected under the level-1 and FP8 gates}
(DeepSeek, held-out English text). Share (\%) of the top lookups by importance, over
all $24$ heads, that both gates select.}
\label{tab:overlap}
\small
\begin{tabular}{lccc}
\toprule
 & Top $10\%$ & Top $23.5\%$ & Top $41.2\%$ \\
\midrule
Block 2 & 98.1 & 98.8 & 99.3 \\
Block 15 & 92.8 & 94.6 & 96.2 \\
\bottomrule
\end{tabular}
\end{table}

\paragraph{Accuracy with quantized gates or values.}
Table~\ref{tab:kvsplit-ci} gives the intervals behind Table~\ref{tab:kvsplit}. In the
\emph{gates only} configuration, the model serves level $k$ of the ladder, so keys and
gates come from it, and every lookup reads its value from the FP8 table; in the
\emph{values only} configuration it serves FP8, and every lookup reads its value from
level $k$. With only the gates quantized, PopQA stays at the level of FP8 at every depth
(changes of $+0.08$ to $+0.15$ points, and no interval end below $-0.22$). With only the
values quantized, the loss is that of quantizing both: the difference between the two
is $-0.16$, $+0.08$, and $-0.04$ points at the three depths, and every interval
contains zero. The predictions that we registered before the runs, $0$ for the gates
and $-4.19$, $-1.01$, and $-0.27$ for the values, are within $0.2$ points of the
measurements.

\begin{table}[h]
\centering
\caption{\textbf{PopQA when only the gates, only the values, or both come from the
quantized table} (DeepSeek). Change from FP8 ($52.39$) in points with the
paired $95\%$ interval over $14{,}267$ questions; the last column is the prediction
registered for the values-only runs.}
\label{tab:kvsplit-ci}
\small
\setlength{\tabcolsep}{4pt}
\begin{tabular}{lcccc}
\toprule
Bits & Gates only & Values only & Both & Predicted (values) \\
\midrule
1.06 & $+0.08$ $[-0.22, +0.37]$ & $-4.03$ $[-4.51, -3.55]$ & $-3.87$ $[-4.35, -3.40]$ & $-4.19$ \\
2.13 & $+0.15$ $[-0.14, +0.43]$ & $-0.81$ $[-1.17, -0.44]$ & $-0.88$ $[-1.24, -0.52]$ & $-1.01$ \\
3.19 & $+0.13$ $[-0.16, +0.42]$ & $-0.19$ $[-0.51, +0.13]$ & $-0.15$ $[-0.47, +0.17]$ & $-0.27$ \\
\bottomrule
\end{tabular}
\end{table}

\subsection{Access Frequency, Gate, and Importance}
\label{app:frequency}

\paragraph{Access counts.}
The counting corpus is skewed. For a $2$-gram head, half of the accesses fall on
about $52{,}000$ rows ($0.33\%$ of the rows), and half of the rows are never
accessed; for a $3$-gram head, half of the accesses fall on $1.56$ million rows, and
for a $4$-gram head on $3.98$ million rows ($25\%$ of the rows).

\paragraph{Frequent rows carry little importance.}
Table~\ref{tab:bands} bins the lookups of the $2$-gram heads by the access count of
their row. On held-out English text, rows seen more than $1{,}000$ times serve
$32\%$ of the lookups and carry $4.5\%$ (block~2) and $1.9\%$ (block~15) of the
importance, whereas rows seen at most $100$ times serve $42\%$ of the lookups and
carry $79\%$ and $88\%$. On PopQA the gap is wider: rows never seen in the counting
corpus serve $14\%$ of the lookups and carry $23\%$ of the importance, and rows seen
more than $1{,}000$ times serve $32\%$ and carry $3.8\%$; if the lookups were weighted
by their gate energy alone, those rows would carry $10.9\%$, so the gate lowers their
share from $32\%$ to $11\%$ and the smaller norms of the rows to $3.8\%$. Over all $24$
heads of block~2, rows never seen serve $6.3\%$ of the lookups and carry $12\%$ of the
importance, and rows seen more than $1{,}000$ times serve $13\%$ and carry $2.0\%$.
Knowledge questions read rarer rows than English text does, and Chinese text reads them
far more often: $43\%$ of the lookups of $2$-gram heads on Chinese text fall on rows
that never occur in the (mostly English) counting corpus.

\begin{table}[h]
\centering
\caption{\textbf{Lookups and importance by access count of the row} ($2$-gram heads
of DeepSeek; percentage of the lookups and of the importance of
\eqref{eq:distortion} in each bin).}
\label{tab:bands}
\small
\begin{tabular}{lrrrrrr}
\toprule
 & \multicolumn{2}{c}{English, block 2} & \multicolumn{2}{c}{English, block 15} & \multicolumn{2}{c}{PopQA, block 2} \\
\cmidrule(lr){2-3}\cmidrule(lr){4-5}\cmidrule(lr){6-7}
Access count & Lookups & Importance & Lookups & Importance & Lookups & Importance \\
\midrule
never seen & 4.2 & 15.0 & 4.2 & 19.5 & 14.2 & 22.7 \\
1 & 3.1 & 9.3 & 3.2 & 12.1 & 9.1 & 12.3 \\
2--10 & 11.6 & 25.1 & 11.5 & 29.5 & 13.7 & 22.9 \\
11--100 & 23.0 & 30.0 & 23.0 & 26.6 & 15.9 & 24.3 \\
101--1{,}000 & 26.1 & 16.1 & 26.1 & 10.4 & 15.2 & 13.9 \\
1{,}001--10{,}000 & 18.5 & 3.6 & 18.5 & 1.7 & 26.2 & 3.4 \\
$>10{,}000$ & 13.5 & 0.9 & 13.5 & 0.3 & 5.6 & 0.4 \\
\bottomrule
\end{tabular}

\end{table}

\paragraph{Gate and access count.}
Table~\ref{tab:spearman} gives the Spearman correlation between the gate and the
access count of the row. It is negative in all $18$ combinations of layer, order, and
corpus, but its size varies from $-0.63$ to $-0.02$: it is strongest for $2$-gram
heads and for code, and weak for $4$-gram heads and for Chinese. Rows seen often
also have smaller norms: the Spearman correlation of the row RMS with the access
count is $-0.45$ for $2$-gram heads, between $-0.27$ and $-0.32$ for $3$-gram heads,
and about $-0.1$ for $4$-gram heads, so both factors of the importance (gate and row
norm) fall with frequency.

\begin{table}[h]
\centering
\caption{\textbf{Spearman correlation between the gate and the access count of the
row} (DeepSeek; the gate is averaged over the four branches; mean over the eight heads
of each $n$-gram order).}
\label{tab:spearman}
\small
\begin{tabular}{llccc}
\toprule
Text & Block & $2$-gram & $3$-gram & $4$-gram \\
\midrule
\multirow{2}{*}{English} & 2 & $-0.51$ & $-0.26$ & $-0.11$ \\
 & 15 & $-0.40$ & $-0.22$ & $-0.09$ \\
\multirow{2}{*}{Chinese} & 2 & $-0.19$ & $-0.04$ & $-0.02$ \\
 & 15 & $-0.18$ & $-0.04$ & $-0.02$ \\
\multirow{2}{*}{Code} & 2 & $-0.63$ & $-0.36$ & $-0.16$ \\
 & 15 & $-0.53$ & $-0.33$ & $-0.15$ \\
PopQA questions & 2 & $-0.40$ & -- & -- \\
\bottomrule
\end{tabular}

\end{table}

\paragraph{Concentration and context.}
Table~\ref{tab:conc} gives the share of the importance carried by the top $1$, $5$,
$10$, and $20\%$ of the lookups. The concentration is high in all three
corpora and in both blocks, and highest in block~15 on PopQA, where the top $1\%$
alone carry $54\%$. The last two columns concern the gate of a single row. The
identity of a row explains much of the variance of its log gate energy (the share
$\eta^2$ among rows looked up at least five times is $0.67$--$0.98$), so an estimate
of a row's gate would help where it can be made. But it can be made only from the
row's own lookups: rows looked up at least $20$ times account for only $9\%$ of the
lookups of held-out English text ($14\%$ on RACE, $68\%$ on PopQA, where prompts
repeat), and even these rows spread: the median ratio between the $90$th and $10$th
percentile of the gate energy of a row's lookups is $2.0$--$2.1$ in block~2 and
$4.3$--$8.8$ in block~15 on the two English corpora.

\begin{table}[h]
\centering
\caption{\textbf{Concentration of importance, and spread of the gate of a row}
(DeepSeek). Share (\%) of the importance of \eqref{eq:distortion} carried by the top
$1$, $5$, $10$, and $20\%$ of the lookups, over all $24$ heads. $\eta^2$: share of the
variance of the log gate energy explained by the row (rows looked up at least five
times). p90/p10: median over rows looked up at least $20$ times of the ratio between
the $90$th and $10$th percentile of the gate energy of their lookups, with the
quartiles in parentheses.}
\label{tab:conc}
\small
\setlength{\tabcolsep}{5pt}
\begin{tabular}{llrrrrcc}
\toprule
Text & Block & Top 1\% & Top 5\% & Top 10\% & Top 20\% & $\eta^2$ & p90/p10 \\
\midrule
\multirow{2}{*}{Held-out English} & 2 & 11 & 32 & 48 & 69 & 0.92 & 2.1 (1.6--2.7) \\
 & 15 & 25 & 53 & 69 & 84 & 0.67 & 8.8 (4.6--15.3) \\
\multirow{2}{*}{RACE passages} & 2 & 12 & 34 & 50 & 69 & 0.92 & 2.0 (1.7--2.4) \\
 & 15 & 12 & 32 & 47 & 66 & 0.71 & 4.3 (3.1--6.4) \\
\multirow{2}{*}{PopQA prompts} & 2 & 10 & 30 & 45 & 65 & 0.98 & 1.2 (1.1--1.4) \\
 & 15 & 54 & 74 & 84 & 92 & 0.95 & 2.0 (1.5--2.8) \\
\bottomrule
\end{tabular}

\end{table}

\providecommand{\method}{GateRQ}

\section{Residual Ladder}
\label{app:ladder}

\paragraph{Quantizers.}
Each code quantizes one level at a time, and a level $k>1$ encodes the residual that
the earlier levels leave, with the same quantizer. \emph{TCQ} \citep{tcq} uses QTIP's
bitshift trellis \citep{qtip} with a $16$-bit state, the computed 3INST codebook, tail
biting, and one FP16 scale per row. \emph{PQ} \citep{pq} divides a row by its FP16
norm, cuts it into $d_e/8$ sub-vectors of eight dimensions, and codes each with one of
$256$ codewords (an $8$-bit index, one bit per element). \emph{INT} rounds to nearest
within groups of $32$ elements that share an E4M3 scale (at one bit the values are
$\pm s/2$); the scale is chosen by a search over clipping values that minimizes the
mean squared error, and the tensor-level coefficient is a power of two. We apply no
rotation: the rows are already close to independent Gaussian vectors
(Appendix~\ref{app:assumptions}), and a rotation brought no gain. vLLM reads DeepSeek's
tables only as FP8 with E8M0 scales, so every level table of DeepSeek is re-encoded in
this format, which is exact for INT and adds $0.063$--$0.065$ percentage points to the
relative error of PQ and TCQ.

\paragraph{Error per level.}
Table~\ref{tab:ladder} gives the relative error
$\varepsilon_k=\|\mathbf e-\hat{\mathbf e}^{(k)}\|_2^2/\|\mathbf e\|_2^2$ after each
level. The first level of TCQ leaves $26.9\%$ of the energy on DeepSeek, the second
$7.2\%$, the third $1.9\%$. Every level removes less error than the one before it,
so the prices $\theta_k=\Delta b_k/\Delta\varepsilon_k$ of \eqref{eq:read-rule} grow
with $k$ (Table~\ref{tab:theta}) and the sets of lookups that read each level are
nested. With the TCQ ladder on DeepSeek, a lookup is worth refining from level~1 to
level~2 when its importance exceeds $5.4\mu$, and from level~2 to level~3 when it
exceeds $20.1\mu$. The errors and the prices of the two models are close.

\begin{table}[h]
\centering
\caption{\textbf{Relative error of the ladder after each level} ($\%$). Bits per level
include the scale.}
\label{tab:ladder}
\small
\setlength{\tabcolsep}{5pt}
\begin{tabular}{lcccccccc}
\toprule
 & \multicolumn{2}{c}{Bits per level} & \multicolumn{3}{c}{DeepSeek} & \multicolumn{3}{c}{Qwen} \\
\cmidrule(lr){2-3}\cmidrule(lr){4-6}\cmidrule(lr){7-9}
Code & DeepSeek & Qwen & Level 1 & Level 2 & Level 3 & Level 1 & Level 2 & Level 3 \\
\midrule
TCQ & 1.0625 & 1.10 & 26.9 & 7.2 & 1.9 & 27.2 & 7.4 & 2.0 \\
PQ & 1.0625 & 1.10 & 32.2 & 10.6 & 3.7 & 32.1 & 10.4 & 3.5 \\
INT & 1.25 & 1.25 & 35.6 & 12.3 & 5.6 & 35.6 & 12.2 & 5.4 \\
\bottomrule
\end{tabular}

\end{table}

\begin{table}[h]
\centering
\caption{\textbf{Price of a level}, $\theta_k=\Delta b_k/\Delta\varepsilon_k$ in bits per
unit of relative error, from Table~\ref{tab:ladder}.}
\label{tab:theta}
\small
\begin{tabular}{lcccc}
\toprule
 & \multicolumn{2}{c}{DeepSeek} & \multicolumn{2}{c}{Qwen} \\
\cmidrule(lr){2-3}\cmidrule(lr){4-5}
Code & $\theta_2$ & $\theta_3$ & $\theta_2$ & $\theta_3$ \\
\midrule
TCQ & 5.4 & 20.1 & 5.6 & 20.4 \\
PQ & 4.9 & 15.4 & 5.1 & 16.0 \\
INT & 5.4 & 18.6 & 5.4 & 18.2 \\
\bottomrule
\end{tabular}

\end{table}

\paragraph{Ladder and direct codes.}
A ladder costs little against a code of the same depth. On DeepSeek, two levels of
TCQ leave $7.2\%$ at $2.13$ bits, against $6.86\%$ at $2.06$ bits for a direct
two-bit code, and three levels leave $1.9\%$ at $3.19$ bits, against $1.77\%$ at
$3.06$ bits for a direct three-bit code; two levels of PQ leave $10.6\%$ against
$9.72\%$, and two levels of INT $12.3\%$ at $2.50$ bits against $10.76\%$ at $2.25$
bits. For a Gaussian source the distortion--rate bound is $2^{-2R}$, which is $25\%$,
$6.25\%$, and $1.56\%$ at $R=1$, $2$, and $3$ bits per element before scales; the TCQ
ladder is $0.3$, $0.6$, and $0.9$~dB above it. Gaussian sources can in principle be
refined without loss \citep{equitz1991}, so the gap of the ladder to the direct code
comes from the quantizers and not from the nesting.

\section{Read Budgets and Bit Accounting}
\label{app:bits}

\paragraph{Bits per lookup.}
We report logical bits per looked-up element: the bits that the chosen depths
require, including the scale of every level read. A level of the TCQ and PQ
ladders costs $1+16/d_e$ bits per element, one bit plus a 16-bit scale per row,
which is $1.0625$ on DeepSeek ($d_e=256$) and $1.10$ on Qwen ($d_e=160$). A level
of the INT ladder costs $1.25$ bits, since it carries an E4M3 scale per $32$
elements. The bytes physically transferred can exceed the logical rate
(Appendix~\ref{app:impl}).

\paragraph{Rate of a read plan.}
If every level costs $b_1$ bits, a plan in which a fraction $s_2$ of the lookups
reads at least level~2 and a fraction $s_3\le s_2$ reads level~3 costs
$b_1(1+s_2+s_3)$ bits per lookup. The fractions of Section~\ref{sec:setup} give
$1.0625\,(1+0.4118)=1.50$ and $1.0625\,(1+0.60+0.3765)=2.10$ bits on DeepSeek,
and $1.10\,(1+0.3636)=1.50$ and $1.10\,(1+0.56+0.349)=2.10$ bits on Qwen. The
INT ladder, with $b_1=1.25$, reads $1.76$ and $2.47$ bits on DeepSeek and
$1.70$ and $2.39$ on Qwen (Table~\ref{tab:plans}). The $1.50$-bit setting fixes $s_2$
at $0.4118$; the $2.10$-bit setting leaves the split between the two deeper levels
open, and we use one split, the same for all rules, codes, and tasks.

\begin{table}[h]
\centering
\caption{\textbf{Read plans} and the bits per lookup that the accuracy runs report
for them (identical for the four rules in every setting).}
\label{tab:plans}
\small
\begin{tabular}{llccccc}
\toprule
 & & \multicolumn{2}{c}{Fractions} & \multicolumn{3}{c}{Bits per lookup} \\
\cmidrule(lr){3-4}\cmidrule(lr){5-7}
Model & Setting & $s_2$ & $s_3$ & TCQ & PQ & INT \\
\midrule
\multirow{2}{*}{DeepSeek} & $1.50$ bits & $0.4118$ & $0$ & $1.500$ & $1.500$ & $1.765$ \\
 & $2.10$ bits & $0.60$ & $0.3765$ & $2.100$ & $2.100$ & $2.471$ \\
\multirow{2}{*}{Qwen} & $1.50$ bits & $0.3636$ & $0$ & $1.500$ & $1.500$ & $1.705$ \\
 & $2.10$ bits & $0.56$ & $0.349$ & $2.100$ & $2.100$ & $2.386$ \\
\bottomrule
\end{tabular}
\end{table}

\paragraph{Bits stored.}
The accuracy tables report the bits read per lookup. \method{} keeps every level that a
lookup may read: two levels ($2.13$ bits on DeepSeek) in the $1.50$-bit setting and
three ($3.19$ bits) in the $2.10$-bit setting. A rule that depends only on the row,
Prior and Frequency, needs the deeper levels only for the rows that it can select,
which is less. The Random rule that we evaluate redraws its choice for every lookup, so
it needs the full ladder like \method{}.

\section{Implementation and Evaluation}
\label{app:impl}

\subsection{Native Implementation}
\label{app:native}

We implement \method{} for DeepSeek-V4.1-Flash as a patch of vLLM~0.30
\citep{pagedattention} that replaces its Engram layer and is loaded when the module is
imported, so the vLLM installation stays unmodified. Level~1 of every row (the $1$-bit
code and its FP16 scale, $34$~bytes) stays in GPU memory, $3.04$~GiB per GPU. Levels~2
and~3 are packed into fixed-size records of $36$ or $68$~bytes and kept in one pinned
host copy per node, which the eight ranks share through \texttt{/dev/shm}: $25.8$~GiB
at $1.50$ bits and $48.6$~GiB at $2.10$ bits, against $188.7$~GiB for FP8 tables
offloaded to host memory. Only DeepSeek has a native implementation; the accuracy
results of both models come from an emulation of the same read rule inside vLLM
(Appendix~\ref{app:protocol}).

In a forward pass, the ranks look up the level-1 codes of their heads and all-gather
them ($36$~bytes per lookup instead of the $512$ bytes of a bf16 row), decode them, and
compute the gates and the importance estimate of \eqref{eq:importance-estimate}. One
Triton kernel computes the quantile thresholds of the batch for up to $16{,}384$ lookups
per layer, and larger batches use a sort. Each rank then copies the records of the
selected lookups of its own heads from host memory into a fixed-size GPU buffer, and the
ranks all-gather these buffers; fixed-size buffers with masks keep the path compatible
with CUDA Graphs. The fused key and value projection is split without copying, so that
deeper levels refine only the value (Algorithm~\ref{alg:forward}), and every rank decodes
the residuals of all heads, so no all-reduce is needed. The serving path is synchronous,
and the bytes physically transferred depend on packing and alignment, so they can exceed
the logical read rate.

Table~\ref{tab:micro} times the table side of one layer on one GPU of a TP8
node, with cold host reads. It includes the all-gathers, the decoding, the thresholds, and
the host reads; the projections, the gates, and the output are the same in all rows and
are not counted. A host read costs about $20$~ns per record whatever its size between
$32$ and $128$~bytes, and level-1 codes make the all-gather $12.7$ times cheaper than for
bf16 rows. At $8{,}192$ tokens, reading the records of each rank's own heads takes $3.0$
($1.50$ bits) and $2.5$ ($2.10$ bits) times less time than FP8 in host memory, and $2.5$
and $2.0$ times less than FP8 in GPU memory (Figure~\ref{fig:load-time}). The exchange has
a fixed latency, so below about $300$--$500$ tokens it is slower, and it is meant for
prefill blocks. The throughput of Appendix~\ref{app:serving} was measured with a simpler
variant in which every rank reads the records of all heads, with identical outputs (rows
\emph{all heads per rank}). In an $8{,}192$-token prefill block of about $0.64$~s, the
table side of the two layers takes $0.4\%$ to $1.4\%$ of the time in the FP8 and
\method{} rows, so these differences do not show in end-to-end throughput.

\begin{table}[h]
\centering
\caption{\textbf{Table-side time of one layer} ($\mu$s per rank, TP8,
cold host reads). Uniform rows keep the listed levels in GPU memory. \method{} rows:
each rank reads the records of its own heads (the variant of Figure~\ref{fig:load-time});
\emph{all heads per rank}: every rank reads the records of all heads (the variant of the
throughput runs). Operator-level timings of one node; the all-gather is not overlapped
with computation.}
\label{tab:micro}
\small
\begin{tabular}{lrrr}
\toprule
 & \multicolumn{3}{c}{Tokens per forward pass} \\
\cmidrule(lr){2-4}
Table configuration & $1{,}024$ & $4{,}096$ & $8{,}192$ \\
\midrule
FP8 in GPU memory & 414 & 1{,}565 & 3{,}144 \\
FP8 in host memory & 498 & 1{,}918 & 3{,}854 \\
Uniform, level 1 & 104 & 207 & 350 \\
Uniform, levels 1--2 & 113 & 325 & 603 \\
Uniform, levels 1--3 & 150 & 461 & 884 \\
\midrule
\method{}, 1.50 bits & 348 & 731 & 1{,}271 \\
\method{}, 2.10 bits & 361 & 857 & 1{,}573 \\
\method{}, 1.50 bits, all heads per rank & 446 & 1{,}287 & 2{,}866 \\
\method{}, 2.10 bits, all heads per rank & 549 & 1{,}691 & 4{,}565 \\
\bottomrule
\end{tabular}
\end{table}

\subsection{Evaluation Protocol}
\label{app:protocol}

We evaluate on the four benchmarks of Table~\ref{tab:benchmarks}, each in full, with the
LM Evaluation Harness \citep{lmeval} $0.4.13$ as a client of a vLLM service: the model's
chat template with thinking mode off, few-shot examples as earlier turns of the
conversation, greedy decoding, and the harness's default seeds. The four tasks run at the
same time, each with $32$ concurrent requests of up to $8$ prompts, and the prompts are
identical across configurations. DeepSeek runs on $8$ GPUs (TP8) and Qwen on
$4$ (TP4), with a maximum length of $131{,}072$ tokens, in eager mode.

\begin{table}[h]
\centering
\caption{\textbf{Benchmarks and scoring.} TriviaQA has $17{,}944$ records of $9{,}960$
questions; CMMLU has $67$ subjects.}
\label{tab:benchmarks}
\small
\setlength{\tabcolsep}{5pt}
\begin{tabular}{lrp{5.6cm}c}
\toprule
Benchmark & Questions & Score & Shots \\
\midrule
PopQA \citep{popqa} & $14{,}267$ & a possible answer occurs in the first line of the reply & 15 \\
TriviaQA \citep{triviaqa} & $9{,}960$ & exact match & 5 \\
NQ-open \citep{nq} & $3{,}610$ & exact match & 5 \\
CMMLU \citep{cmmlu} & $11{,}582$ & accuracy of the answer letter, by log-likelihood & 5 \\
\bottomrule
\end{tabular}
\end{table}

The accuracy runs serve the model whose tables hold level~1, and a hook acts in every
Engram forward pass: it computes the gates from the served rows, gives every lookup a
depth by the rule of the configuration and the quotas of the batch
(Appendix~\ref{app:bits}), reads the selected rows from tables that hold level~$k$, and
replaces the shared value of the positions by the projection of the mixed rows, while
keys, and thus gates, stay those of level~1. At the first forward pass of each layer the
hook checks its output against that of the model's kernel, and it counts the rows read
per level. The rules rank the lookups of a batch by $\hat I$ of
\eqref{eq:importance-estimate} (Gate); by the same score with the gate replaced by the
expected gate energy of the row (Prior), tabulated per head and per access-count bin on
held-out text; by the access count of the row (Frequency); or by a random number
(Random). Ties are broken at random, and Random draws afresh for every lookup. A forward
pass holds at most $4{,}096$ tokens on DeepSeek ($3.9$ thousand on average) and $7.7$
thousand on average on Qwen, so the lookups of many requests share the quotas of one
batch.

Differences between configurations are paired by question. We report $95\%$ percentile
intervals of a paired bootstrap over questions, with $10{,}000$ resamples ($4{,}000$ for
the comparisons by popularity quartile); TriviaQA, whose $17{,}944$ records come from
$9{,}960$ questions, is resampled by question. Each configuration was run once, so the
intervals cover the sampling of questions, not the variation between runs. The fraction of
the gap recovered in Section~\ref{sec:results-ablation} is
$(\text{rule}-\text{level }1)/(\text{levels }1\text{--}2-\text{level }1)$ in PopQA
accuracy, with its interval from the same resamples.

The offline proxy of Section~\ref{sec:results-ablation} computes the importance-weighted
error $\bar\varepsilon=\sum I_{t,j}\varepsilon_{\ell_{t,j}}/\sum I_{t,j}$ of a read plan on
the records of PopQA prompts and converts it into a predicted PopQA loss by
$a\bar\varepsilon+b\bar\varepsilon^{3}$. The coefficients of each model were fitted once
to earlier end-to-end runs (uniform codes, mixtures of rows, halved sets of hash heads, and
the zero table); the leave-one-out error on the quantized runs is $0.24$ points on DeepSeek
and $0.09$ on Qwen. We wrote the predicted change of every configuration to a file before
each queue of runs; Appendix~\ref{app:predictions} compares the predictions with the
measurements.

\providecommand{\method}{GateRQ}

\section{Extended Results}
\label{app:results}

\subsection{INT and PQ Ladders}
\label{app:fourtask}

Table~\ref{tab:main-alt} gives PopQA and the mean of the four tasks for the INT and PQ
ladders; the TCQ ladder is in Table~\ref{tab:main}. In all $12$ settings (two models,
three codes, and two read budgets), \method{} has the best PopQA accuracy and the best
four-task average of the four rules.

\begin{table}[t]
\centering
\caption{\textbf{Results with the INT and PQ ladders.} As Table~\ref{tab:main}, with
PopQA and the mean of the four tasks. \textbf{INT}: every level an INT1 code (groups of
$32$ with E4M3 scales, $1.25$ bits) at the same read shares as TCQ, so each lookup reads
$1.18\times$ (DeepSeek) and $1.14\times$ (Qwen) the bits of TCQ. \textbf{PQ}: every level a PQ code
($8$-dimensional sub-vectors, $256$ codewords, and a per-row norm; $1.06$ bits on
DeepSeek, $1.10$ on Qwen) at the same read shares and bits per lookup as TCQ. The PopQA
half-widths of the paired $95\%$ intervals against FP8 are $0.32$--$0.52$ points for INT
and $0.30$--$0.51$ for PQ.}
\label{tab:main-alt}
\footnotesize
\setlength{\tabcolsep}{3.5pt}
\begin{tabular}{lrrrrrrr}
\toprule
 & \multicolumn{3}{c}{INT} & \multicolumn{3}{c}{PQ} \\
\cmidrule(lr){2-4}\cmidrule(lr){5-7}
Method & Read & PopQA & Avg. & Read & PopQA & Avg. \\
\midrule
\multicolumn{7}{l}{\textit{DeepSeek-V4.1-Flash}} \\
FP8 & 8.25 & 52.39 & 64.34 & 8.25 & 52.39 & 64.34 \\
\midrule
Uniform, level~1 & 1.25 & 46.79 & 62.35 & 1.06 & 46.81 & 62.17 \\
Uniform, levels~1--2 & 2.50 & 50.56 & 63.58 & 2.13 & 51.39 & 63.88 \\
Uniform, levels~1--3 & 3.75 & 51.85 & 64.12 & 3.19 & 51.87 & 64.36 \\
\cmidrule(lr){1-7}
\multicolumn{7}{l}{\textit{INT: $1.76$, PQ: $1.50$ bits per lookup}} \\
\ours Gate (\method{}) & 1.76 & \textbf{50.52} & \textbf{63.64} & 1.50 & \textbf{51.35} & \textbf{63.87} \\
Prior & 1.76 & 49.33 & 63.19 & 1.50 & 50.25 & 63.54 \\
Random & 1.76 & 48.71 & 62.97 & 1.50 & 49.16 & 63.03 \\
Frequency & 1.76 & 48.34 & 62.88 & 1.50 & 48.49 & 62.95 \\
\cmidrule(lr){1-7}
\multicolumn{7}{l}{\textit{INT: $2.47$, PQ: $2.10$ bits per lookup}} \\
\ours Gate (\method{}) & 2.47 & \textbf{51.61} & \textbf{63.99} & 2.10 & \textbf{52.07} & \textbf{64.29} \\
Prior & 2.47 & 50.50 & 63.67 & 2.10 & 51.05 & 63.89 \\
Random & 2.47 & 49.84 & 63.40 & 2.10 & 50.41 & 63.67 \\
Frequency & 2.47 & 49.44 & 63.21 & 2.10 & 49.72 & 63.37 \\
\midrule
\multicolumn{7}{l}{\textit{Qwen3.8-Flash-Next}} \\
FP8 & 8.00 & 39.55 & 59.18 & 8.00 & 39.55 & 59.18 \\
\midrule
Uniform, level~1 & 1.25 & 35.80 & 57.69 & 1.10 & 36.59 & 58.00 \\
Uniform, levels~1--2 & 2.50 & 38.28 & 58.87 & 2.20 & 38.63 & 58.90 \\
Uniform, levels~1--3 & 3.75 & 39.26 & 59.11 & 3.30 & 39.31 & 59.06 \\
\cmidrule(lr){1-7}
\multicolumn{7}{l}{\textit{INT: $1.70$, PQ: $1.50$ bits per lookup}} \\
\ours Gate (\method{}) & 1.70 & \textbf{37.91} & \textbf{58.71} & 1.50 & \textbf{38.29} & \textbf{58.76} \\
Prior & 1.70 & 37.32 & 58.39 & 1.50 & 37.90 & 58.63 \\
Random & 1.70 & 36.85 & 58.15 & 1.50 & 37.16 & 58.32 \\
Frequency & 1.70 & 36.48 & 57.87 & 1.50 & 36.73 & 58.06 \\
\cmidrule(lr){1-7}
\multicolumn{7}{l}{\textit{INT: $2.39$, PQ: $2.10$ bits per lookup}} \\
\ours Gate (\method{}) & 2.39 & \textbf{38.89} & \textbf{59.06} & 2.10 & \textbf{39.28} & \textbf{59.07} \\
Prior & 2.39 & 38.30 & 58.76 & 2.10 & 38.56 & 58.85 \\
Random & 2.39 & 37.61 & 58.45 & 2.10 & 38.05 & 58.59 \\
Frequency & 2.39 & 37.02 & 58.21 & 2.10 & 37.47 & 58.31 \\
\bottomrule
\end{tabular}

\end{table}

\subsection{Gate Against the Selection Rules}
\label{app:rules}

\paragraph{PopQA.}
Table~\ref{tab:rules} gives the paired differences behind Table~\ref{tab:main} and
Figure~\ref{fig:accuracy-popqa}, for all three codes. \method{}
beats Prior, Random, and Frequency in all $36$ comparisons, by $0.39$ to $2.86$ points,
and all intervals exclude zero. Reading the next level of the uniform ladder instead of
using \method{} is better by $0.34$--$0.36$ points in $3$ of the $12$ settings (Qwen
with INT at both budgets and with PQ at $1.50$ bits), at $0.70$--$1.36$ bits more per
lookup, and not significantly different in the other nine.

\begin{table}[h]
\centering
\caption{\textbf{PopQA: \method{} against the selection rules, and against reading the
next level of the uniform ladder.} Paired differences in points with $95\%$ intervals.
Columns 3--5: \method{} minus the rule (positive: \method{} is better). Column 6:
uniform levels 1--2 (at $1.50$ bits) or 1--3 (at $2.10$ bits) minus \method{} (positive:
the uniform read is better). Bold: the interval excludes $0$. Last column: bits per
lookup that \method{} reads fewer than the uniform read.}
\label{tab:rules}
\scriptsize
\setlength{\tabcolsep}{2.5pt}
\begin{tabular}{llccccr}
\toprule
Setting & Bits & Prior & Random & Frequency & Uniform & Saved \\
\midrule
DeepSeek TCQ & 1.50 & {\boldmath $1.14$ $[0.76, 1.51]$} & {\boldmath $1.62$ $[1.22, 2.03]$} & {\boldmath $1.84$ $[1.44, 2.23]$} & $+0.13$ $[-0.18, +0.44]$ & 0.63 \\
DeepSeek TCQ & 2.10 & {\boldmath $0.70$ $[0.34, 1.07]$} & {\boldmath $1.43$ $[1.05, 1.82]$} & {\boldmath $1.87$ $[1.46, 2.29]$} & $+0.06$ $[-0.27, +0.37]$ & 1.09 \\
DeepSeek INT & 1.50 & {\boldmath $1.18$ $[0.80, 1.58]$} & {\boldmath $1.81$ $[1.37, 2.24]$} & {\boldmath $2.17$ $[1.75, 2.60]$} & $+0.04$ $[-0.29, +0.37]$ & 0.74 \\
DeepSeek INT & 2.10 & {\boldmath $1.11$ $[0.72, 1.50]$} & {\boldmath $1.77$ $[1.34, 2.19]$} & {\boldmath $2.17$ $[1.75, 2.60]$} & $+0.24$ $[-0.07, +0.55]$ & 1.28 \\
DeepSeek PQ & 1.50 & {\boldmath $1.10$ $[0.71, 1.49]$} & {\boldmath $2.19$ $[1.77, 2.61]$} & {\boldmath $2.86$ $[2.44, 3.29]$} & $+0.04$ $[-0.27, +0.36]$ & 0.63 \\
DeepSeek PQ & 2.10 & {\boldmath $1.02$ $[0.65, 1.39]$} & {\boldmath $1.66$ $[1.25, 2.07]$} & {\boldmath $2.35$ $[1.93, 2.77]$} & $-0.20$ $[-0.51, +0.11]$ & 1.09 \\
Qwen TCQ & 1.50 & {\boldmath $0.44$ $[0.13, 0.76]$} & {\boldmath $0.79$ $[0.43, 1.14]$} & {\boldmath $1.55$ $[1.17, 1.91]$} & $+0.16$ $[-0.11, +0.43]$ & 0.70 \\
Qwen TCQ & 2.10 & {\boldmath $0.71$ $[0.39, 1.04]$} & {\boldmath $1.06$ $[0.70, 1.40]$} & {\boldmath $1.51$ $[1.14, 1.87]$} & $+0.18$ $[-0.10, +0.44]$ & 1.20 \\
Qwen INT & 1.50 & {\boldmath $0.59$ $[0.26, 0.91]$} & {\boldmath $1.06$ $[0.67, 1.45]$} & {\boldmath $1.43$ $[1.02, 1.84]$} & {\boldmath $+0.36$ $[+0.09, +0.64]$} & 0.80 \\
Qwen INT & 2.10 & {\boldmath $0.60$ $[0.26, 0.93]$} & {\boldmath $1.28$ $[0.90, 1.68]$} & {\boldmath $1.88$ $[1.47, 2.28]$} & {\boldmath $+0.36$ $[+0.10, +0.64]$} & 1.36 \\
Qwen PQ & 1.50 & {\boldmath $0.39$ $[0.06, 0.72]$} & {\boldmath $1.14$ $[0.75, 1.51]$} & {\boldmath $1.56$ $[1.17, 1.96]$} & {\boldmath $+0.34$ $[+0.06, +0.62]$} & 0.70 \\
Qwen PQ & 2.10 & {\boldmath $0.72$ $[0.39, 1.06]$} & {\boldmath $1.23$ $[0.86, 1.59]$} & {\boldmath $1.81$ $[1.43, 2.19]$} & $+0.04$ $[-0.23, +0.29]$ & 1.20 \\
\bottomrule
\end{tabular}

\end{table}

\paragraph{Other tasks.}
Table~\ref{tab:tasks} counts, for each task, how often \method{} is significantly better
than, not significantly different from, or significantly worse than a rule over the $12$
settings. On TriviaQA and CMMLU, \method{} is significantly better in $20$ and $21$ of
the $36$ comparisons; Frequency is significantly worse than \method{} in all $12$
settings on CMMLU. On NQ-open the rules are hard to separate: \method{} is significantly
better in one comparison. The one significant exception is TriviaQA with TCQ at $2.10$
bits on DeepSeek, where Frequency is $0.35$ points better.

\begin{table}[h]
\centering
\caption{\textbf{\method{} against each rule on the four tasks.} Over the $12$ settings:
number of comparisons in which \method{} is significantly better / not significantly
different / significantly worse (paired $95\%$ intervals). Last column: mean advantage
of \method{} over the $36$ comparisons, in points.}
\label{tab:tasks}
\small
\begin{tabular}{lccccc}
\toprule
Task & Prior & Random & Frequency & All & Mean \\
\midrule
PopQA & 12/0/0 & 12/0/0 & 12/0/0 & 36/0/0 & $+1.38$ \\
TriviaQA & 5/7/0 & 8/4/0 & 7/4/1 & 20/15/1 & $+0.33$ \\
CMMLU & 1/11/0 & 8/4/0 & 12/0/0 & 21/15/0 & $+0.26$ \\
NQ-open & 0/12/0 & 0/12/0 & 1/11/0 & 1/35/0 & $+0.16$ \\
\bottomrule
\end{tabular}

\end{table}

\subsection{Registered Predictions}
\label{app:predictions}

Table~\ref{tab:pred} lists the predictions of the offline proxy for the change in PopQA
accuracy, written to files before the runs, next to the measurements. $57$ of the $66$
predictions lie within $0.4$ points of the measurement: $21$ of $22$ for TCQ, and $18$ of
$22$ for each of INT and PQ, whose largest deviations are $0.66$ and $0.67$ points. The
proxy predicts the order of the four rules correctly in all $12$ settings. It errs
conservatively on DeepSeek, where the measured change is on average $0.22$ points less
negative than predicted, and by $0.09$ points the other way on Qwen; five of the nine
deviations above $0.4$ are for Frequency.

\begin{table}[h]
\centering
\caption{\textbf{Predicted and measured PopQA change from FP8} (points).
For each code, the prediction of the offline proxy, registered before the run, and the
measurement.}
\label{tab:pred}
\small
\setlength{\tabcolsep}{4pt}
\begin{tabular}{lcccccc}
\toprule
 & \multicolumn{2}{c}{TCQ} & \multicolumn{2}{c}{INT} & \multicolumn{2}{c}{PQ} \\
\cmidrule(lr){2-3}\cmidrule(lr){4-5}\cmidrule(lr){6-7}
Configuration & Predicted & Measured & Predicted & Measured & Predicted & Measured \\
\midrule
\multicolumn{7}{l}{\textit{DeepSeek-V4.1-Flash}} \\
Uniform, level 1 & $-4.19$ & $-3.87$ & $-5.99$ & $-5.59$ & $-5.25$ & $-5.58$ \\
Uniform, levels 1--2 & $-1.01$ & $-0.88$ & $-1.76$ & $-1.83$ & $-1.50$ & $-1.00$ \\
Uniform, levels 1--3 & $-0.27$ & $-0.15$ & $-0.79$ & $-0.54$ & $-0.51$ & $-0.51$ \\
\cmidrule(lr){1-7}
\multicolumn{7}{l}{\textit{1.50 bits per lookup}} \\
Gate & $-1.20$ & $-1.02$ & $-1.99$ & $-1.87$ & $-1.71$ & $-1.04$ \\
Prior & $-1.87$ & $-2.15$ & $-2.85$ & $-3.06$ & $-2.48$ & $-2.14$ \\
Random & $-2.77$ & $-2.64$ & $-4.03$ & $-3.68$ & $-3.53$ & $-3.23$ \\
Frequency & $-3.27$ & $-2.85$ & $-4.71$ & $-4.04$ & $-4.13$ & $-3.90$ \\
\cmidrule(lr){1-7}
\multicolumn{7}{l}{\textit{2.10 bits per lookup}} \\
Gate & $-0.40$ & $-0.20$ & $-0.95$ & $-0.78$ & $-0.66$ & $-0.32$ \\
Prior & $-0.99$ & $-0.90$ & $-1.68$ & $-1.89$ & $-1.36$ & $-1.34$ \\
Random & $-1.87$ & $-1.63$ & $-2.82$ & $-2.54$ & $-2.42$ & $-1.98$ \\
Frequency & $-2.43$ & $-2.07$ & $-3.56$ & $-2.95$ & $-3.09$ & $-2.66$ \\
\midrule
\multicolumn{7}{l}{\textit{Qwen3.8-Flash-Next}} \\
Uniform, level 1 & $-2.47$ & $-2.50$ & $-3.43$ & $-3.76$ & $-2.99$ & $-2.96$ \\
Uniform, levels 1--2 & $-0.62$ & $-0.67$ & $-1.05$ & $-1.28$ & $-0.88$ & $-0.92$ \\
Uniform, levels 1--3 & $-0.17$ & $0.00$ & $-0.44$ & $-0.29$ & $-0.29$ & $-0.24$ \\
\cmidrule(lr){1-7}
\multicolumn{7}{l}{\textit{1.50 bits per lookup}} \\
Gate & $-0.79$ & $-0.83$ & $-1.32$ & $-1.64$ & $-1.11$ & $-1.26$ \\
Prior & $-1.33$ & $-1.27$ & $-2.01$ & $-2.23$ & $-1.69$ & $-1.65$ \\
Random & $-1.77$ & $-1.61$ & $-2.54$ & $-2.70$ & $-2.18$ & $-2.40$ \\
Frequency & $-2.00$ & $-2.38$ & $-2.81$ & $-3.07$ & $-2.44$ & $-2.82$ \\
\cmidrule(lr){1-7}
\multicolumn{7}{l}{\textit{2.10 bits per lookup}} \\
Gate & $-0.28$ & $-0.18$ & $-0.65$ & $-0.66$ & $-0.44$ & $-0.27$ \\
Prior & $-0.82$ & $-0.89$ & $-1.35$ & $-1.25$ & $-1.07$ & $-1.00$ \\
Random & $-1.24$ & $-1.23$ & $-1.84$ & $-1.94$ & $-1.54$ & $-1.51$ \\
Frequency & $-1.36$ & $-1.68$ & $-1.99$ & $-2.54$ & $-1.69$ & $-2.08$ \\
\bottomrule
\end{tabular}

\end{table}

\section{Serving Throughput}
\label{app:serving}

Table~\ref{tab:efficiency} reports native synchronous serving on $8$
GPUs. Compared with FP8 tables in GPU memory, \method{} reduces table
memory from $23.6$ to $3.04$~GiB per GPU, enlarges the KV-cache capacity by
about $8.1\times$ (as reported by vLLM), and improves long-context throughput
by $11\%$--$31\%$. Compared with FP8 host offloading, throughput is similar at
$1.50$--$2.10$ bits per lookup while pinned host memory falls from $188.7$ to
$25.8$--$48.6$~GiB. The gain over GPU-resident FP8 is from the larger KV cache,
not from reading fewer bits: level~1 alone and the two \method{} budgets sit on
the same throughput plateau as FP8 host offloading. These are workload-specific
measurements from single runs. Native four-task accuracy remains to be
cross-checked against the accuracy runs.

Both workloads use $32$K input tokens; they differ in the output length ($1$K
and $2$K tokens) and in the number of concurrent requests ($64$ and $128$).
Every cell is one \texttt{vllm bench serve} run with its own random prompts, so
the prefix cache is not hit. The rows ran on three nodes of the same type; the
$1.50$-bit setting, which ran on all three, differs by at most $0.4\%$ and
$0.8\%$ between them at the two workloads.

\begin{table}[h]
\centering
\caption{\textbf{Native DeepSeek serving (TP8).} GPU: table GiB
per GPU; Host: pinned GiB per node; throughput: output tok/s at $64$ concurrent
requests with $1$K output tokens and at $128$ concurrent requests with $2$K
output tokens ($32$K input tokens in both). Uniform rows keep the listed levels
in GPU memory.}
\label{tab:efficiency}
\small
\setlength{\tabcolsep}{5pt}
\begin{tabular}{lrrrr}
\toprule
 & & & \multicolumn{2}{c}{Throughput} \\
\cmidrule(lr){4-5}
Table configuration & GPU & Host & 64 requests & 128 requests \\
\midrule
FP8 in GPU memory & 23.6 & 0 & 228.3 & 349.6 \\
FP8 host offloading & 0 & 188.7 & 253.3 & 460.1 \\
Uniform, level 1 & 3.04 & 0 & 256.0 & 464.5 \\
Uniform, levels 1--2 & 6.1 & 0 & 256.8 & 464.0 \\
Uniform, levels 1--3 & 9.1 & 0 & 256.4 & 465.0 \\
\method{}, 1.50-bit setting & 3.04 & 25.8 & 254.0 & 458.6 \\
\method{}, 2.10-bit setting & 3.04 & 48.6 & 254.6 & 458.1 \\
\bottomrule
\end{tabular}
\end{table}

\end{document}